\documentclass[twoside,11pt]{article}

\usepackage[abbrvbib, preprint]{jmlr2e}

\usepackage{color,amsmath,stmaryrd,float,mathrsfs}
\usepackage[linesnumbered,ruled]{algorithm2e}

\SetKwInput{KwInput}{Input}                
\SetKwInput{KwOutput}{Output} 

\newtheorem{assumption}{Assumption}

\def\R{\mathbb{R}}
\def\Y{\mathbb{Y}}

\newcommand{\argmin}[1]{ \underset{#1}{\textup{argmin}} }

\usepackage{lastpage}

\ShortHeadings{Model-free generalized fiducial inference}{Williams}
\firstpageno{1}

\begin{document}

\title{Model-free generalized fiducial inference}

\author{\name Jonathan P Williams \email jwilli27@ncsu.edu \\
       \addr Department of Statistics\\
       North Carolina State University\\
       Raleigh, NC, USA}

\editor{}

\maketitle

\begin{abstract}
Conformal prediction (CP) was developed to provide finite-sample probabilistic prediction guarantees.  While CP algorithms are a relatively general-purpose approach to uncertainty quantification, with finite-sample guarantees, they lack versatility.  Namely, the CP approach does not {\em prescribe} how to quantify the degree to which a data set provides evidence in support of (or against) an arbitrary event from a general class of events.  In this paper, tools are offered from imprecise probability theory to build a formal connection between CP and generalized fiducial (GF) inference.  These new insights establish a more general inferential lens from which CP can be understood, and demonstrate the pragmatism of fiducial ideas.  The formal connection establishes a context in which epistemically-derived GF probability matches aleatoric/frequentist probability.  Beyond this fact, it is illustrated how tools from imprecise probability theory, namely lower and upper probability functions, can be applied in the context of the imprecise GF distribution to provide posterior-like, prescriptive inference that is not possible within the CP framework alone.  In addition to the primary CP generalization that is contributed, fundamental connections are synthesized between this new model-free GF and three other areas of contemporary research: nonparametric predictive inference (NPI), conformal predictive systems/distributions, and inferential models (IMs).  
\end{abstract}


\section{Introduction}\label{sec:intro}

The rate at which machine learning algorithms are being developed and advanced for high-stakes applications is greatly exceeding the pace at which safe and reliable methods for quantifying their uncertainty are becoming understood.  Moreover, it is largely undetermined how to properly quantify uncertainty in their performance guarantees, and there is no generally accepted standard of accountability of stated uncertainties.  As discussed in \cite{shafer2021}---among many other references---a trouble with non-frequentist interpretations of probability are their practical limitations for verifiability.  Frequentist interpretations of probability yield explicit definitions based on probabilistic statements that can be tested and verified (if only through theoretical simulation), and admit tangible attributes of data models, such as {\em validity} of predictions (e.g., control over type 1 error rates).  Notions of validity are fundamental to developing methods and procedures that have any chance at being reliable when applied in the context of uncertainty quantification (for prediction and inference, alike).  It is for these reasons that statisticians must afford a high premium to repeated sampling properties.  CP was developed to provide finite-sample probabilistic prediction guarantees by leveraging the calibration inherent in applying the empirical hold-out method for training/testing a machine learning prediction rule \citep{vovk2005}.  Growing momentum for applications and developments of CP has occurred in recent years.

While CP algorithms are a relatively general-purpose approach to uncertainty quantification, with finite-sample guarantees, they lack versatility.  Namely, the CP approach does not {\em prescribe} how to quantify the degree to which a data set provides evidence in support of (or against) an arbitrary event from a general class of events.  For instance, within the Bayesian paradigm, the degree to which a data set provides evidence in support of (or against) an event is quantified by the posterior probability of the event, {\em for any measurable event}.  Bayesian inference, however, operates by the usual Kolmogorov axioms for probability calculus, and is thereby subject to the false confidence theorem \citep{balch2019, martin2019, carmichael2018}, rendering it provably unreliable.  The false confidence theorem is mathematical justification for the fact that precise probabilistic-based statistical inferences (e.g., those based on posterior probabilities) are provably unreliable in the sense that there always exists a false hypothesis (with positive Lebesgue measure) having arbitrarily large epistemic (e.g., posterior) probability, with arbitrarily large aleatory (i.e., frequency/frequentist) probability.  This theorem arose to explain a troubling phenomenon occurring in the Bayesian analysis of satellite trajectory data \citep{balch2019}.  Consequences of the false confidence theorem can, however, be avoided via imprecise probability calculus \citep{martin2019}.

In this paper, tools are offered from imprecise probability theory to build a formal connection between CP and GF inference \citep{hannig2016}.  These new insights establish a more general inferential lens from which CP can be understood, and demonstrates the pragmatism of fiducial ideas.  The key observation about the CP framework is that the rank of a nonconformity score actually defines a {\em data generating association} with an auxiliary discrete-uniform distribution.  It is proved in this article that applying the GF inference framework to this rank-based data generating association leads to a model-free approach---which is hereafter refer to as {\em model-free GF}---for constructing GF prediction sets, from which CP sets arise.  This also means that the epistemically-derived GF probability matches the aleatoric/frequentist probability on CP sets.  Beyond this fact, it is illustrated how tools from imprecise probability theory, namely lower and upper probability functions, can be applied in the context of the imprecise GF distribution to provide posterior-like, prescriptive inference that is not possible within the CP framework alone.  Furthermore, beyond the primary CP generalization that is contributed, fundamental connections are synthesized between this new model-free GF and three other areas of contemporary statistics research: NPI, conformal predictive systems/distributions, and IMs.

First, the model-free GF framework turns out to be a generalization of the NPI approach \citep[e.g.,][]{coolen1998,augustin2004,coolen2024} built from the Dempster-Hill procedure/assumption \citep{dempster1963,hill1968,hill1988}.  In fact, the Dempster-Hill assumption is satisfied trivially and more generally within the model-free GF paradigm, and under this assumption non-asymptotic, sub-exponential concentration inequalities are derived to establish root-$n$ consistency, around the true distribution of the data, of every probability measure in the credal set of the imprecise model-free GF distribution.  

Second, conformal predictive distributions \cite[e.g.,][]{vovk2019,vovk2022,vovk2024} are also an extension of NPI ideas but with the aim of constructing predictive probability distribution functions from monotonic nonconformity scores.  This approach is advertised as fiducial in the sense of constructing a probability distribution from a pivot, but is more aligned with the confidence distribution characterization of fiducial ideas \citep[i.e.,][]{xie2013,schweder2016} rather than the GF or IM characterizations.  In contrast, the model-free GF characterization does not rely on monotonicity of nonconformity scores, nor does it lead to a precise probabilistic predictive distribution.  In fact, the model-free GF imprecise probabilistic characterization is more aligned with the NPI prescriptions for inference, while still leveraging CP calibration constructions beyond the Dempster-Hill procedure/assumption.

Third, the data generating association implemented to construct the model-free GF distribution by way of the GF {\em switching principle} \citep{hannig2016} can alternatively be implemented by way of a {\em predictive random set} to construct an IM \citep{martin2015} for predictive inference, as in \cite{cella2022}.  Similar to the model-free GF construction, this IM construction inherently elicits the formulation of CP sets.  The difference between the two constructions is that the focal sets of the imprecise IM distribution {\em are} the nested CP sets, whereas, the focal sets of the imprecise model-free GF distribution are the disjoint set-differences of the nested CP sets.  It turns out that the disjoint nature of the focal sets, as in the model-free GF construction, is necessary for the statistical consistency of the probability measures in the credal set.  The IM predictive distribution in \cite{cella2022} was designed instead to achieve an imprecise probabilistic notion of finite-sample frequentist validity, a property endowed by {\em consonance} \citep{shafer1976,cella2022}; the consonant IM construction is similarly established using the model-free GF framework in Section \ref{sec:IM}, thereby also establishing imprecise probabilistic validity.

Next, because precise probabilistic approximations from the credal set associated with model-free GF belief/plausibility functions may be desirable, it is illustrated in this article how one such approximation can be arrived at based on the principle of maximum entropy.  This should be consistent with the pignistic transformation based on the generalized insufficient reason principle advocated for in \cite{smets1990}.  A mapping from a credal set to a probability distribution is often referred to as {\em a pignistic transformation}; the terminology being that beliefs are held at the credal level while decisions are made at the pignistic level \citep{smets1990}.  The problem of constructing pignistic transformations has been considered, more broadly, in the approximate reasoning research community.  For example, \cite{dubois2004} provides conditions and theorems for transformation between possibility/necessity measures and probability measures.  Still, no general theory for such constructions or evaluations of such mappings exists \citep{grunwald2018}, and so there remain many open research questions concerning precise probability approximations of a credal set, especially from a statistical inference perspective.    

Imprecise probabilities, and in particular, the roles of [non-additive] belief and plausibility functions have been extensively developed within the context of Dempster-Shafer (DS) theory \citep[][]{dempster1966,shafer1976}.  DS theory has seen varied applications, namely in artificial intelligence communities \citep[e.g.,][]{bloch1996,vasseur1999,denoeux2000,basir2007,denoeux2008,diaz-mas2010}, but has largely not been applied in mainstream statistical literatures.  The lack of attention from the statistics communities may be partly attributed to major barriers to computation \citep{shafer2021}.  Remarkably, a recent solution drawing positive attention has been provided in the article \cite{jacob2021} for the computation of DS inference on categorical data, a problem that had been open for 55 years.  Regardless of the success/failure of the DS theory for inference, the related ideas developed for lower and upper probabilities in \cite{shafer1976} are very useful and apply more broadly, as demonstrated with the IM framework.  In particular, the utility of a {\em don't know} category has hugely important implications on statistical inference, as demonstrated/discussed in \cite{balch2019, martin2019, carmichael2018, williams2021}.

Another line of work related to the model-free GF formulation of CP sets is the development of {\em risk controlling prediction sets} \citep{bates2021,angelopoulos2023,angelopoulos2025}.  In the risk controlling prediction set framework CP ideas are generalized to provide guarantees on the expected loss between the response label and CP set, where the specification of the loss function is arbitrary (based on minimal assumptions).  In principle, the model-free GF ideas could also be applied to exchangeable instances of loss functions, but the imprecision of the model-free GF formulation more naturally lends itself to the analysis of {\em upper/lower previsions} of a loss (i.e., supremum/infimum of the expectation of the loss function, over a credal set of probability measures).  Indeed, the risk controlling prediction set perspective inspired investigation of the latter in \cite{williams2024decision}.

The remainder of this paper is organized as follows.  Section \ref{sec:safety} lays out a general discussion and motivating remarks on various notions of safety and reliability in prediction and inference; namely, such notions go far beyond the unconditional validity emphasized in this article.  The imprecise probabilistic phenomenon of {\em dilation} is also offered for interpretation.  Section \ref{sec:cp_from_gf} serves to introduce the fundamental ideas for CP and the GF inference paradigm, followed by construction of the framework being proposed for {\em model-free GF inference}.  The imprecise nature of this construction is examined in Section \ref{sec:imprecision}, including connections to NPI, conformal predictive distributions, and IMs.  An illustration is offered in Section \ref{sec:pignistic_illustration} for the construction of a pignistic transformation from the model-free GF credal set to a [precise] probability distribution, followed by a brief survey of other such transformations implicitly considered in the GF literature.  Concluding remarks are provided in Section \ref{sec:conclusion}, and the Julia programming language codes to reproduce all numerical illustrations and figures are publicly available at: \url{https://jonathanpw.github.io/research.html}.

\section{Safe probability, testing, and inference}\label{sec:safety}

The developments in this article are governed by a particular notion of reliability, namely finite-sample, unconditional, type 1 control of prediction error; commonly this is referred to as a type of {\em unconditional validity}.  Consider the following motivating remarks.  

At the American Society of Clinical Oncology conference in Chicago in June 2022 \citep{tie2022}, it was discussed that a new liquid biopsy can help identify the need for adjuvant therapy in stage II colon cancer thereby potentially avoiding post-operative chemotherapy, which for colon cancer, can cause peripheral neuropathy.  Suppose a machine learning algorithm is trained to identify the need for adjuvant therapy with 95\% confidence reported.  How is this confidence defined?  Is it defined as the reported error on a test set?  Is it a Bayesian posterior probability?  Is it some sort of averaging over a collection of predictions?  All of these are widely accepted notions of {\em confidence}, but they all represent different quantifications of uncertainty with varying (if any) guarantees for how the algorithm might perform on future data.  When a meteorologist predicts that there is 95\% chance of rain tomorrow, it might not be so problematic to not understand in what sense {\em 95\% chance} is reliable or verifiable (if at all), but when an algorithm says there is 95\% chance that a person does not need post-operative chemotherapy with potentially life debilitating side effects, there are serious ramifications for how to interpret that quantification of uncertainty.  Moreover, for such quantification of uncertainty to be reliable there must be a mechanism for accountability.  Finite-sample, unconditional, type 1 error control facilitates such accountability in that 95\% confidence is defined unambiguously by the algorithmic attribute that in no more than 5\% of instances of rejecting the hypothesis `post-operative chemotherapy is needed' will post-operative chemotherapy actually be needed---an error rate that can be verified on real data of any sample size.

CP and model-free GF are reliable in this sense of unconditional validity for uncertainty quantification of predictions.  For statistical inference of a population parameter IMs and, more recently, universal inference \citep{wasserman2020,park2023} guarantee the same type of unconditional validity, as does generalized universal inference \citep{dey2024} for a risk minimizer; see also \cite{dey2023} on valid inference for machine learning model parameters.  Unconditional validity has limitations, though.  For instance, suppose that of colon cancer patients, 99\% are of age 45 years or older, and that the type 1 error rate of rejecting `post-operative chemotherapy is needed' is 4\% for such patients but 100\% for patients younger than 45.  Unconditionally, the error rate is 4.96\% in this example, so the algorithm is reliable with respect to a stated guarantee of 5\% error rate, but it is clearly not safe for young colon cancer patients.  A stronger notion of {\em conditional validity}, e.g., finite-sample type 1 error control conditioned on covariate values, is required, but not readily achievable within the CP nor model-free GF constructions, unless further information is available.  Note that this is more general than the special case of all meaningful values of a covariate being known a-priori, in which case Mondrian CP \citep[][and analogously model-free GF]{vovk2005,hjort2025} could be applied (essentially training independently on each covariate-valued class).  

Beyond unconditional and conditional validity, the article \cite{grunwald2018} builds an entire theory of various notions of reliability, therein termed ``safety'' and hence the title of this section.  Loosely, in the order of strength of the guarantees offered, these notions are: asymptotic validity, unbiased, confidence safety, pivotal safety, calibration, squared-error optimality, unconditional validity, conditional validity.  Calibration of an algorithm meaning that an event actually occurs in $p\cdot100\%$ of observed instances, for events that are predicted to have probability $p$ by the algorithm, e.g., rainfall occurring on approximately 30\% of days for which the algorithm predicted `probability of rain is .3'.  More recently, notions of reliability have been introduced and developed on in the context of sequential testing problems, termed {\em anytime-validity}; i.e., reliability in the sense of finite-sample type 1 error rate control under {\em optional continuation} or {\em optional stopping} of statistical testing regimes determined before or after data are observed.  These properties are achieved via the construction of {\em e-values} and {\em e-processes}, by leveraging Markov's inequality and properties of non-negative supermartingales such as Ville's inequality.  Furthermore, whereas p-values generally cannot be combined (e.g., across studies) to produce a p-value, e-values can, e.g., the average of e-values is an e-value.  See the following references for a current review on anytime-validity and related topics: \cite{wasserman2020,ramdas2023,grunwald2024,grunwald2024a}.

One final remark on the topic of reliability is in order.  An important issue is raised about the phenomenon of {\em dilation} in \cite{grunwald2018} concerning conditional validity as it relates to imprecise probability; see also \cite{seidenfeld1993}.  Consider the following simple illustration modified from \cite{grunwald2018}.  Suppose it is of interest to predict the event $\{U = 1\}$ for some random variable $U \sim \text{Bernoulli}(.4)$.  In this case $P(U = 1) = .4$.  Now assume further that $U$ depends in some unknown way on some other random variable $V \sim \text{Bernoulli}(.5)$ that is observed.  Without knowing how $U$ depends on $V$, an imprecise probabilistic assessment might model the uncertainty by including all probability distributions $P$ on $\{0,1\}\times\{0,1\}$ that satisfy:
\[
.4 = P(U = 1) = P(U = 1 \mid V = 0) \cdot .5 + P(U = 1 \mid V = 1) \cdot .5.
\]
 Consequently, the imprecise probabilistic assessment yields $P(U = 1 \mid V = v) \in [0,.8]$ regardless of the observed value $v \in \{0,1\}$.  This means that after having gained information about $V$ we actually know less about the event $\{U = 1\}$ than we did ignoring $V$, i.e., information about $\{U = 1\}$ is lost.  This phenomenon of dilation is indeed perplexing because the obvious remedy of ignoring $V$ conflicts with the precept that information should never be useless.  I offer, however, the reconciliation that while information should never be useless, we must know how to use it for it to be useful; otherwise, it should not be used, as in the illustration.  

Of course, imprecise probabilities are not necessarily a consequence of information lost due to dilation.  And the developments to follow in this article illustrate how imprecise probabilities are a natural consequence of frequentist calibration.

\section{Constructing CP sets from GF inference}\label{sec:cp_from_gf}

Throughout this text the notion of a population parameter will be used to refer to unknown population quantities of interest.  Depending on the context, a parameter may take an arbitrary value.  Most common examples of parameters are objects described in, but not limited to, scalar-, vector-, or matrix-value form.  Though, a population parameter of interest may also be defined as an infinite-dimensional object such as a distribution function, for example.  In the case of prediction, the unknown parameter value is the datum value to be predicted.  For a random sample $y_{1}, \dots, y_{n}$, of size $n$, denote $y_{n+1}$ as the datum value to be predicted, and assume that these values are, respectively, realizations of the random variables $Y_{1}, \dots, Y_{n}, Y_{n+1} \overset{\text{iid}}{\sim} Y$, where $Y$ represents the random variable from a population model.  Moving forward, the shorthand, $y \sim Y$ is taken to mean $y$ is an observed instance of the random variable $Y$.

Classical statistical inference on the unknown value of $y_{n+1}$ would be to assume a parametric model for $Y$ and construct prediction sets either inspired by large sample theory (e.g., an asymptotic confidence interval) or from a Bayesian posterior predictive distribution.  While both approaches are considered reasonable, they both allow the practitioner to avoid accountability to a stated nominal level of confidence.  Without knowledge of the population model, the parametric prediction sets are not guaranteed to achieve their nominal coverage, at least non-asymptotically, and Bayesian posterior credible sets are not promised to be calibrated to any notion of reliability.  This is highly problematic because without rigorous justification of a stated nominal level of confidence, practitioners can claim any level without consequence, and so it is not clear in what sense {\em probabilities} are meaningful.  We need for probabilities assigned to prediction sets to be inherently meaningful in a manner that is mathematically or empirically verifiable, and so we must begin by defining the properties they ought to have.  Such is the fundamental principle of the CP approach, discussed next.

For any a-priori fixed $\alpha \in (0,1)$, suppose that $\Gamma^{\alpha}_{n}$ is an $\alpha$ level prediction set for $y_{n+1}$, constructed from observed data $y_{1}, \dots, y_{n}$.  Next, assuming $y_{n+1} \sim Y$, let $\xi$ be the binary indicator of the event that $\Gamma^{\alpha}_{n}$ does {\em not} contain $y_{n+1}$, and take $\xi_{1},\xi_{2},\dots$ to represent a sequence of independent, repeated samples of $\xi$.  Then $\Gamma^{\alpha}_{n}$ is said to be (conservatively) {\em valid} if $\xi_{1},\xi_{2},\dots$ is dominated in distribution by that of a sequence of iid $\alpha$ weighted coin tosses \citep{vovk2005}.  This notion is stated more concisely in Definition \ref{validity_type1}.  
\begin{definition}[Type 1 validity -- \cite{cella2022}]\label{validity_type1}
Let $\{\Gamma^{\alpha}_{n} \, : \, \alpha \in (0,1)\}$ be a family of prediction sets constructed from observed data $y_{1}, \dots, y_{n} \sim Y$, and assume $y_{n+1} \sim Y$.  Denoting by $P$ the probability measure associated with $Y$, the family of prediction sets is type 1 valid if, for all $(\alpha, n, P)$,
$
P\big(\Gamma^{\alpha}_{n} \ni Y_{n+1}\big) \ge 1 - \alpha.
$
\end{definition}
Attributable to an emphasis on bounding type 1 errors, conservative validity is often simply referred to as validity.  It turns out that CP sets are valid in this sense, for finite sample sizes, as discussed next.

\subsection{Conformal prediction}\label{sec:cp}

The basic principle for any CP set is that it is constructed from an algorithm providing finite-sample guarantees, called a {\em conformal algorithm} and stated here as Algorithm \ref{alg:cp_alg}.  Perhaps the simplest context for introducing a conformal algorithm is the classification scenario where {\em exchangeable} examples $y_{1},\dots,y_{n} \sim Y$ are observed, as in Definition \ref{exchangeability}, and the task is to determine whether some new value $y$ is exchangeable with $y_{1}, \dots, y_{n}$.  Note that exchangeability of data is a slightly weaker condition than assuming iid data.

The CP strategy is to first define a measure of {\em nonconformity}, $\Psi \, : \, \lbag\R^{n}\rbag\times\R \to \R$, such that $\Psi(y^{n+1}_{-i},y_{i})$, for $i \in \{1,\dots,n+1\}$, is a meaningful measure of how different the value $y_{i}$ is from the values $y_{1},\dots,y_{i-1},y_{i+1},\dots,y_{n+1}$, where $y^{n+1}_{-i} := \{y_{1},\dots,y_{n+1}\} \setminus \{y_{i}\}$.  Then the assertion that $y_{n+1}$ is exchangeable with $y_{1}, \dots, y_{n}$ is dismissed if the value $\Psi(y^{n+1}_{-(n+1)},y_{n+1})$ falls in the $\alpha$ tail region of the empirical distribution of the values $\Psi(y^{n+1}_{-i},y_{i})$, for $i \in \{1,\dots,n+1\}$.  When the context is clear, for conciseness let $t_{i}(y_{i}) := \Psi(y^{n+1}_{-i},y_{i})$, for $i \in \{1,\dots,n+1\}$.

Using the conformal algorithm, a CP set denoted by $\Gamma_{n}^{\alpha}$ is constructed as the set of all $y$ such that the conformal algorithm returns value 1.  As exhibited by Theorem \ref{theorem:cp_cons_valid}, the novelty of the conformal algorithm is its finite-sample control of type 1 errors at the stated nominal level $\alpha$, for any user-specified level $\alpha \in (0,1)$, and it is sufficient to only assume exchangeability of the data examples (i.e., a model-free assumption).  In fact, the exchangeability of the data is not necessary so long as $t_{1}(Y_{1}), \dots, t_{n+1}(Y_{n+1})$ are exchangeable.

\begin{definition}[Exchangeability]\label{exchangeability} 
A sequence $Y_{1}, Y_{2}, \dots$ with probability measure $P$ is said to be exchangeable if for every integer $n > 0$, every permutation $\sigma$ on $\{1,\dots,n\}$, and every $P$-measurable set $E$, 
$
P\big\{(Y_{1}, \dots, Y_{n}) \in E\big\} = P\big\{(Y_{\sigma(1)}, \dots, Y_{\sigma(n)}) \in E\big\}.
$
\end{definition}

\medskip
\begin{algorithm}[H]\label{alg:cp_alg}
\KwInput{Nonconformity measure $\Psi : \lbag\R^{n}\rbag\times\R \to \R$, measurable; exchangeable examples $y_{1},\dots,y_{n}$; an arbitrary value $y$; and significance level $\alpha \in (0,1)$.}
\KwOutput{Logical value; 1 indicates that $y_{1},\dots,y_{n},y$ are exchangeable, and 0 else.}
Denote $y_{n+1} := y$\;
\For{$i \in \{1, \dots, n+1\}$}
{
Compute $t_{i}(y_{i}) = \Psi(y^{n+1}_{-i},y_{i})$\;
}
Set $p_{n}(y_{n+1}) := \frac{1}{n+1}\sum_{i=1}^{n+1}1\{t_{i}(y_{i}) \ge t_{n+1}(y_{n+1})\}$\;
\Return{ $1\{p_{n}(y_{n+1}) > \alpha\}$\; }
\caption{Conformal algorithm \citep{vovk2005}}
\end{algorithm}
\medskip

\begin{theorem}[\cite{vovk2005}]\label{theorem:cp_cons_valid}
If the random variables $Y_{1},\dots,Y_{n+1} \sim Y$ are exchangeable, then a CP set is [type 1] valid, as in Definition \ref{validity_type1}.
\end{theorem}
{\noindent \bf Proof.}  This result is established in \cite{vovk2005}, but I provide an alternative explicit proof in the Appendix.  The proof follows by first observing that a type 1 error is the event $\{p_{n}(y_{n+1}) \le \alpha\}$, and then showing that $P\{p_{n}(Y_{n+1}) \le \alpha\} \le \alpha$.\hfill $\blacksquare\\$

While provably valid, the CP approach lacks the versatility to assign confidence to assertions $\{y_{n+1} \, : \, B \ni y_{n+1}\}$ if $B$ does not coincide with a CP set at some level.  In Section \ref{sec:mfgf}, a model-free formulation of GF inference is constructed that is able to assign GF-based probability the same as the level of a CP set (and is thus valid), but is also able to assign imprecise probabilities (i.e., lower and upper probabilities) to all other assertions.  The necessary requisites on GF inference are introduced in the next section.  
\begin{definition}\label{def:p-value_fun}
The function $p_{n}(y) := \frac{1}{n+1}\sum_{i=1}^{n+1}1\{t_{i}(y_{i}) \ge t_{n+1}(y)\}$, as in Algorithm \ref{alg:cp_alg}, will be referred to as the `conformal p-value function'.
\end{definition}

\subsection{GF inference}\label{sec:gf_intro}

The motivating assumption for GF inference is an explicit association between data $Y$ and an auxiliary variable $U$ through some deterministic function $G$ that depends on unknown population parameters of interest, $\theta$.  Expressed as,
\begin{equation}\label{dge}
Y = G(U,\theta),
\end{equation}
the association is typically referred to as a {\em data generating equation}.  A key aspect of the assumption is that the auxiliary variable has a completely known and fully specified distribution.  The auxiliary variable can be understood similar to the notion of a pivotal quantity that might be constructed in the context of statistical testing or bootstrapping.  The goal is to build epistemic inference on the unknown $\theta$ by using the assumption of association (\ref{dge}).  

From the GF inference perspective, the association (\ref{dge}) represents a mapping from a parameter space $\Theta$ to the support $\Y$ of the datum $Y$, and as such, once data are observed, an inverse mapping would contain valuable information about the unknown value $\theta$.  More precisely, given an observed data set $y_{1},\dots,y_{n}$ generated independently from (\ref{dge}) there necessarily exists a corresponding set of auxiliary variable values $u_{1}, \dots, u_{n}$ such that the unknown value $\theta$ solves the system of equations,
$y_{1} = G(u_{1},\theta), \cdots, y_{n}  = G(u_{n},\theta)$.
If the set of auxiliary values $u_{1}, \dots, u_{n}$ were known, then this would be a deterministic inverse problem.  Nonetheless, although $u_{1}, \dots, u_{n}$ are unknown, it is assumed that the set comprises values that were generated independently and identically from the assumed known and fully specified distribution of the auxiliary variable $U$.    

Accordingly, these facts motivate the formal definition of a GF distribution of $\theta$, presented next in Definition \ref{gfd_definition}.  This definition is interpreted as: the probability that $\theta$ is contained in a set $B$ is equal to the probability, under the distribution of $U_{1},\dots,U_{n}$, that the limiting quantity is equal to a $\theta$ that lies in the set $B$.  Note that in this definition, $y_{1}, \dots, y_{n}$ are regarded as fixed while $U_{1}, \dots, U_{n}$ are random.  Thus, the GF distribution is a distributional statistic for the unknown value $\theta$, inheriting its uncertainty from the distribution of the auxiliary random variable, same as $Y_{1}, \dots, Y_{n}$.  In \cite{hannig2016} this is referred to as the {\em switching principle}.  The notion of a distributional statistic for a fixed but unknown parameter, i.e., $\theta$, is analogous to the role played by the posterior distribution in the Bayesian framework.
\begin{definition}[\cite{hannig2016}]\label{gfd_definition}
Given an observed data set $y_{1}, \dots, y_{n}$ generated independently from (\ref{dge}), a GF distribution on a parameter space $\Theta$ is defined as the weak limit,
\[
\lim_{\epsilon\to0}\bigg\{ \argmin{\vartheta \in \Theta}\sum_{i=1}^{n}\|y_{i} - G(U_{i},\vartheta)\|^{2} \ \Big| \ \min_{\vartheta \in \Theta}\sum_{i=1}^{n}\|y_{i} - G(U_{i},\vartheta)\|^{2} \le \epsilon \bigg\},
\]
where $G$ is a deterministic function, and the distribution of $U_{1}, \dots, U_{n}$ is fully known and specified.
\end{definition}

For discrete-valued data, the limit $\epsilon \to 0$ in Definition \ref{gfd_definition} reduces to setting $\epsilon = 0$ leading to an imprecise probability distribution over $\Theta$.  For example, in the case of binomial$(m,\theta)$ data, the data generating equation (\ref{dge}) may take the form:
\[
Y = \sum_{k=1}^{m}1\{U_{k} < \theta\},
\]
where $U_{1},\dots,U_{m} \overset{\text{iid}}{\sim} \text{uniform}(0,1)$.  For an observed instance $y$ from this data generating equation, the GF distribution for $\theta$ is obtained by replacing the unobserved $u_{1},\dots,u_{m}$ that generated $y$ with an independent copy of the auxiliary variables $U_{1}^{\star},\dots,U_{m}^{\star} \overset{\text{iid}}{\sim} \text{uniform}(0,1)$, and setting $\epsilon = 0$ in Definition \ref{gfd_definition}.  This leads to the imprecise GF distribution for $\theta$ defined by the interval-valued random variable of the form
$(U_{(y)}^{\star},U_{(y+1)}^{\star}] \subseteq \Theta$,
where $U_{1}^{\star},\dots,U_{m}^{\star} \overset{\text{iid}}{\sim} \text{uniform}(0,1)$ and $U_{(k)}^{\star}$ denotes the $k$-th order statistic of $U_{1}^{\star},\dots,U_{m}^{\star}$; i.e., $U_{(y)}^{\star}$ is a random variable characterizing the uncertainty about $u_{(y)}$---the unobserved auxiliary value corresponding to the observed data value $y$.  

\subsection{Model-free GF inference}\label{sec:mfgf}

The GF inference approach begins with the assumption that data is generated independently from a data generating equation as in (\ref{dge}).  Such an assumption is model-based, and requires explicit knowledge of the deterministic function $G$ in (\ref{dge}) along with the distribution of the auxiliary variables.  Instead, consider Assumption \ref{cont_exchange}.

\begin{assumption}\label{cont_exchange}
The variables $Y_{1},\dots,Y_{n+1} \in \Y$ are exchangeable and continuous (in the sense of having a continuous distribution function with respect to the Lebesgue measure).
\end{assumption}

\begin{figure}[H]
\centering
\includegraphics[scale=.5, trim={0mm 0mm 0mm 75mm}, clip]{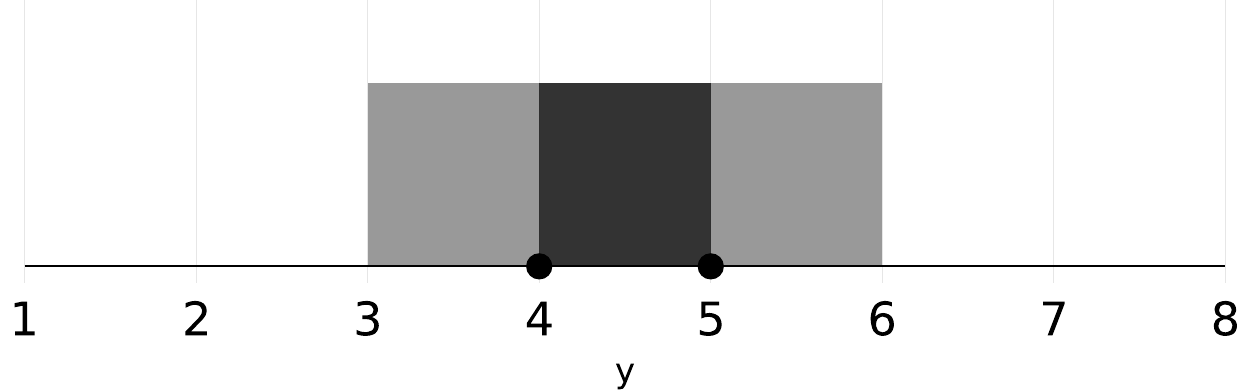}
\caption{\footnotesize Hypothetical observed univariate data with $y_{1} = 4$, $y_{2} = 5$, and $n = 2$.  With nonconformity measure $\Psi(y^{n+1}_{-i},y_{i}) := |\text{mean}(y_{-i}^{n+1}) - y_{i}|$, the inner black region represents the values of $y_{n+1}$ that would have rank 1 (i.e., $A_{n}(1) = \big\{ y \, : \, \text{rank}[t_{n+1}(y)] = 1 \big\}$), the outer grey region represents the values of $y_{n+1} \in A_{n}(2) = \big\{ y \, : \, \text{rank}[t_{n+1}(y)] = 2 \big\}$, and the outermost white region represents the values of $y_{n+1} \in A_{n}(n+1) = \big\{ y \, : \, \text{rank}[t_{n+1}(y)] = n+1 \big\}$.}\label{fig:gf_regions}
\end{figure}

If an injective nonconformity measure $\Psi$ can be constructed for the data, then under Assumption \ref{cont_exchange} a model-free data generating {\em association} for $Y_{n+1}$ is given by:
\begin{equation}\label{model_free_dge}
\text{rank}\{t_{n+1}(Y_{n+1})\} = V \sim \text{uniform}\{1,\dots, n+1\},
\end{equation}
where $\text{rank}\{t_{n+1}(Y_{n+1})\}$ denotes the position or {\em rank} of $t_{n+1}(Y_{n+1})$ in the order statistics (in ascending order) of the sample $t_{1}(Y_{1}),\dots,t_{n+1}(Y_{n+1})$; i.e.,
$$
\text{rank}(t_{j}(y_{j})) := 1 + \sum_{i=1}^{n+1}1\{t_{j}(y_{j}) > t_{i}(y_{i})\},
$$
for $j \in \{1,\dots,n+1\}$.  In this model-free approach, the phrase data generating {\em association} is used in place of data generating {\em equation} because knowledge of the true auxiliary variable value in equation (\ref{model_free_dge}) does {\em not} fully determine the datum value $y_{n+1}$.  Nonetheless, the GF inference algorithm can be applied with reference to the datum variable $\text{rank}\{t_{n+1}(Y_{n+1})\}$, as usual, but for inference on $y_{n+1}$.  First, replace the unobserved true auxiliary variable in (\ref{model_free_dge}) with an independent copy, $V^{\star} \sim \text{uniform}\{1,\dots, n+1\}$.  Second, apply the switching principle to obtain an imprecise GF distribution of the to-be-predicted value $y_{n+1}$ as a distribution over the random {\em focal sets},
\begin{equation}\label{imprecise_gf}
A_{n}(V^{\star}) := \argmin{y\in\Y}\big\{|\text{rank}(t_{n+1}(y)) - V^{\star}|\big\} = \big\{ y \, : \, \text{rank}(t_{n+1}(y)) = V^{\star} \big\},
\end{equation}
as illustrated in Figure \ref{fig:gf_regions}.  
\begin{remark}
The continuity requirement of Assumption \ref{cont_exchange} together with the injectivity assumption about $\Psi$ are to ensure that $t_{i}(Y_{i}) \ne t_{j}(Y_{j})$ a.s.~for any $i \ne j$.  Otherwise, equation (\ref{model_free_dge}) is misspecified because the support of the random variable $\text{rank}\{t_{n+1}(Y_{n+1})\}$ may not include the entire set $\{1,\dots, n+1\}$.
\end{remark}
The imprecise GF probability measure denoted by $\mu : 2^\Y \to [0,1]$ is defined only over the focal sets by
\begin{equation}\label{eq:focal_pt_prob}
\mu\{A_{n}(V^{\star})\} = \pi^{n}_{v}\bigg( V^{\star} = 1 + \sum_{i=1}^{n+1}1\{t_{n+1}(y_{n+1}) > t_{i}(y_{i})\} \bigg) = \frac{1}{n+1},
\end{equation}
where $\pi^{n}_{v}$ denotes the discrete uniform probability mass function associated with the auxiliary variable $V^{\star}$.

\begin{figure}[H]
\centering
\includegraphics[scale=.45, trim={0mm 0mm 0mm 0mm}, clip]{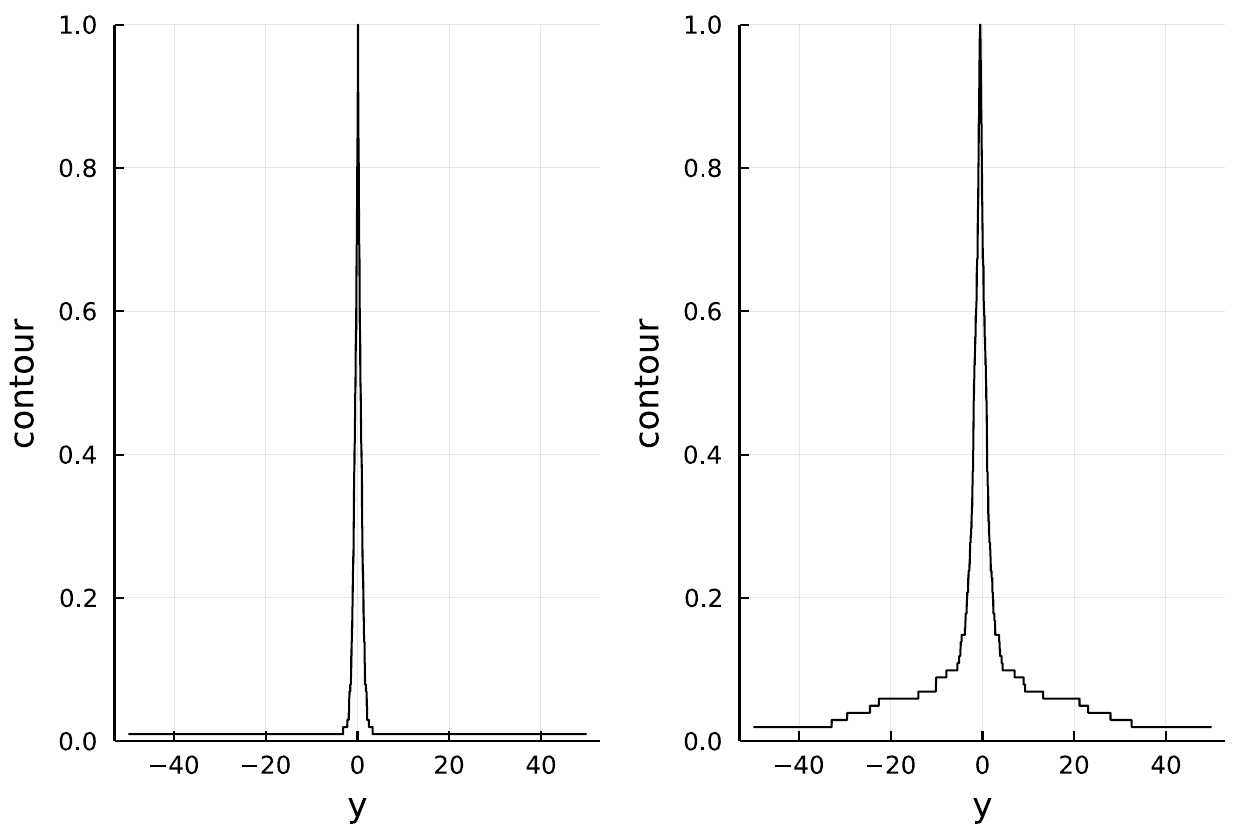}
\caption{\footnotesize Both panels display plots of the GF contour, $f_{n}(y) = \mu\big\{ \Omega_{n}(V^{\star}) \ni y \big\}$; the left and right plots are based on samples of $n = 100$ realizations from the standard Gaussian and standard Cauchy distributions, respectively.}\label{fig:contour_plot}
\end{figure}

An important insight from Figure \ref{fig:gf_regions} is that the imprecise GF distribution of $y_{n+1}$ assigns, in particular, $1/(n+1)$ probability to the outermost region (beyond where any data were observed), and as such, it assigns $n/(n+1)$ probability to the complementary set (within which all of the data were observed).  Moreover, a $k/(n+1)$ GF prediction set is given by,
\begin{equation}\label{eq:omega}
\Omega_{n}(k) := \bigcup_{1\le v \le k}A_{n}(v) = \big\{ y \, : \, \text{rank}[t_{n+1}(y)] \le k \big\},
\end{equation}
for any $k \in \{1,\dots,n+1\}$.  Theorem \ref{gf_contour_valid}, below, relates this model-free GF prediction set to a CP set via the {\em GF contour function}: 
\begin{equation}\label{eq:gf_contour}
f_{n}(y) := \mu\big\{ \Omega_{n}(V^{\star}) \ni y \big\}.
\end{equation}
As a simple example, for the hypothetical data displayed in Figure \ref{fig:gf_regions}, the GF contour is
\[
f_{n}(y) = 
\begin{cases}
1 & \text{ if } \ y \in (4,5) \\
\frac{2}{3} & \text{ if } \ y \in (3,4]\cup[5,6) \\
\frac{1}{3} & \text{ else }
\end{cases}.
\]
More interesting examples of GF contours, for synthetic Gaussian and Cauchy data, are plotted in Figure \ref{fig:contour_plot}.  The important implication of Theorem \ref{gf_contour_valid} is that $\Upsilon_{n}^{\alpha} := \{ y : f_{n}(y) > \alpha\}$ is a [type 1] valid, model-free GF prediction set, as stated in Corollary \ref{gf_prediction_valid}. 

At any level $\alpha \in (0,1)$, the region $\Upsilon_{n}^{\alpha}$ is easily determined in the plots in Figure \ref{fig:contour_plot} by drawing a horizontal line at the value of $\alpha$ and including all values of $y$ that satisfy $f_{n}(y) > \alpha$ (i.e., $\Upsilon_{n}^{\alpha}$ is a pre-image set of $f_{n}$).  Although a contour is {\em not} understood as a density function, construction of $\Upsilon_{n}^{\alpha}$ is akin to the construction of high posterior density credible sets in Bayesian inference.

\begin{theorem}\label{gf_contour_valid}
Under Assumption \ref{cont_exchange} the GF contour $f_{n}(y)$ is a conformal p-value function.
\end{theorem}
{\noindent \bf Proof.}  
This result is simply a statement of the fact that using $f_{n}(y_{n+1})$ in place of $p_{n}(y_{n+1})$ in Algorithm \ref{alg:cp_alg} defines a conformal algorithm.  This follows because
\begin{equation}\label{eq:gf_contour_derived}
\begin{split}
f_{n}(y_{n+1}) & = \mu\big\{ \Omega_{n}(V^{\star}) \ni y_{n+1} \big\} \\
& = \mu\big\{ \text{rank}\{t_{n+1}(y_{n+1})\} \le V^{\star} \big\} \\
& = \frac{n+1 - \text{rank}\{t_{n+1}(y_{n+1})\} + 1}{n+1} \\
& = \frac{n+1 - 1 - \sum_{i=1}^{n+1}1\{t_{n+1}(y_{n+1}) > t_{i}(y_{i})\} + 1}{n+1} \\
& = \frac{\sum_{i=1}^{n+1}1\{t_{n+1}(y_{n+1}) \le t_{i}(y_{i})\} }{n+1}. \\
\end{split}
\end{equation}
\hfill $\blacksquare$\\ 

\begin{corollary}\label{gf_prediction_valid}
Under Assumption \ref{cont_exchange} the GF prediction set $\Upsilon_{n}^{\alpha}$ is [type 1] valid, as in Definition \ref{validity_type1}.
\end{corollary}
{\noindent \bf Proof.}  As shown in Theorem \ref{gf_contour_valid}, $f_{n}(y_{n+1})$ is equivalent to $p_{n}(y_{n+1})$ in Algorithm \ref{alg:cp_alg}, and so 
\[
P( \Upsilon_{n}^{\alpha} \not\ni Y_{n+1}) = P\big(f_{n}(Y_{n+1}) \le \alpha \big) \le \alpha, 
\]
as a direct consequence of Theorem \ref{theorem:cp_cons_valid}.
\hfill $\blacksquare$\\

It is clear from Theorem \ref{gf_contour_valid} that CP sets can be constructed from the GF inference paradigm, namely $\Upsilon_{n}^{\alpha}$ is a CP set.  Furthermore, it follows from expression (\ref{eq:gf_contour_derived}) that any CP set as constructed from Algorithm \ref{alg:cp_alg} can be understood as a union of focal sets from the imprecise GF probability distribution of $y_{n+1}$ defined by (\ref{imprecise_gf}).  This fact establishes the fundamental connection between GF inference and CP.

\section{Characterizing the imprecision in the model-free GF distribution}\label{sec:imprecision}

The imprecision in the GF distribution is that the probability mass $\mu$ is only defined for sets of values $A_{n}(v) \subseteq \Y$, for $v \in \{1,\dots,n+1\}$, rather than a mass or density function defined for all points in $\Y$, as in the precise probability scenario; and the imprecision comes from the fact that nothing is being assumed about the underlying distribution of the data.  The novelty of the approach is that the GF framework is, nonetheless, versatile enough to provide inferences and construct CP sets, based on the model-free association (\ref{model_free_dge}) with the sole assumption of exchangeable, continuous data.  Inferences can be facilitated by the construction of belief and plausibility functions, denoted $\underline{\mu}$ and $\overline{\mu}$, respectively, so that for any event $B \subseteq \Y$ pertaining to the prediction of $y_{n+1}$, i.e., $\{y_{n+1} \, : \, B \ni y_{n+1}\}$,
\begin{align*}
\overline{\mu}(B) & := \sum_{j=1}^{n+1}\mu\{A_{n}(j)\}\cdot 1\big\{A_{n}(j)\cap B \ne \emptyset\big\}, \text{ and} \\
\underline{\mu}(B) & := \sum_{j=1}^{n+1}\mu\{A_{n}(j)\}\cdot 1\big\{A_{n}(j)\subseteq B\big\}.
\end{align*}

A probability measure $\Delta$ is naturally considered {\em compatible} with $\mu$ if for every measurable set $B$, 
$
\underline{\mu}(B) \le \Delta(B) \le \overline{\mu}(B).
$
The set of all such probability measures is called the {\em credal set} of $\mu$, and can be expressed as
\[
\mathscr{C}(\mu) := \big\{\Delta \, : \, \Delta(B) \le \overline{\mu}(B), \text{ for any measurable set } B \big\}.
\]
Further, this construction implies that any probability measure $\Delta \in \mathscr{C}(\mu)$ must assign the same probability mass as $\mu$ to any focal set of $\mu$.  This fact is established as a direct consequence of Lemma \ref{lemma:focalpts}. 
\begin{lemma}\label{lemma:focalpts}
A probability measure $\Delta \in \mathscr{C}(\mu)$ if and only if $\Delta\{A_{n}(v)\} = \frac{1}{n+1} = \mu\{A_{n}(v)\}$, for every $v \in \{1,\dots,n+1\}$.
\end{lemma}
{\noindent \bf Proof.} First, suppose that $\Delta \in \mathscr{C}(\mu)$.  Then for any $v \in \{1,\dots,n+1\}$, using the fact that $A_{n}(1),\dots,A_{n}(n+1)$ are mutually disjoint,
\begin{align*}
\mu\{A_{n}(v)\} & = \sum_{j=1}^{n+1}\mu\{A_{n}(j)\}1\big\{A_{n}(j)\subseteq A_{n}(v)\big\} \\
& = \underline{\mu}\{A_{n}(v)\} \\
& \le \Delta\{A_{n}(v)\} \\
& \le \overline{\mu}\{A_{n}(v)\} \\
& = \sum_{j=1}^{n+1}\mu\{A_{n}(j)\}1\big\{A_{n}(j)\cap A_{n}(v) \ne \emptyset\big\} \\
& = \mu\{A_{n}(v)\} 
\end{align*}
The desired result follows by equation (\ref{eq:focal_pt_prob}).

For the converse direction, assume that $\Delta\{A_{n}(v)\} = \frac{1}{n+1}$, for every $v \in \{1,\dots,n+1\}$.  Then, for any measurable set $B$, using the fact that $A_{n}(1),\dots,A_{n}(n+1)$ are mutually disjoint and collectively exhaustive over $\Y$,
\begin{align*}
\Delta(B) & = \sum_{v=1}^{n+1}\Delta\big\{B\cap A_{n}(v)\big\} \\
& \le \sum_{v=1}^{n+1}\Delta\{A_{n}(v)\} 1\big\{A_{n}(v)\cap B\ne \emptyset\big\} \\
& = \sum_{v=1}^{n+1}\frac{1}{n+1} 1\big\{A_{n}(v)\cap B\ne \emptyset\big\} \\
& = \sum_{v=1}^{n+1}\mu\{A_{n}(v)\}1\big\{A_{n}(v)\cap B\ne \emptyset\big\} \\
& = \overline{\mu}(B).
\end{align*}
\hfill $\blacksquare$\\ 

The implication of interest of Lemma \ref{lemma:focalpts} is that any probability measure compatible with the imprecise GF probability measure $\mu$ must assign uniform (i.e., $\frac{1}{n+1}$) probability to each of the mutually disjoint and collectively exhaustive regions $A_{n}(1),\dots,A_{n}(n+1)$.  Assuming a density function exists, the probability density associated with any $\Delta \in \mathscr{C}(\mu)$, however, can have arbitrary shape over each region $A_{n}(v)$, subject to the constraint that it integrates to $\frac{1}{n+1}$.

\subsection{Connection to nonparametric predictive inference}

A special case of the model-free GF inference construction is the NPI construction developed by Coolen and his collaborators since the late 1990s \citep[e.g.,][]{coolen1998,augustin2004,coolen2024}, and the so-called ``Dempster-Hill procedure'' based on the seminal papers \cite{dempster1963,hill1968,hill1988}.  The name {\em Dempster-Hill procedure} is coined in \cite{vovk2024} and is described as the statement \citep[paraphrased from][]{vovk2024}:
\begin{quote}
Given data $y_{1}, \dots, y_{n}$, the probability that the next observation $y$ falls in $(y_{(i)},y_{(i+1)})$ is $1/(n+1)$, for each $i \in \{0,\dots,n\}$; by definition, $y_{(0)} := -\infty$ and $y_{(n+1)} := \infty$.
\end{quote} 
This idea, taken to be the fundamental assumption of NPI, is argued in \cite{vovk2024} to trace back as far as \cite{jeffreys1932}.  Taking the nonconformity score to be each datum itself, i.e., $t(Y_{i}) := Y_{i}$, and deriving the model-free GF distribution exactly yields the Dempster-Hill procedure, by virtue of equation (\ref{eq:focal_pt_prob}).  From here, all of NPI follows via the same imprecise probability calculus of lower and upper probabilities described above \citep{coolen2024}, but is generalized by model-free GF for arbitrary choices of nonconformity score.

The Dempster-Hill procedure is used in \cite{vovk2024} to construct the ``Dempster-Hill pivot'', a continuity-corrected conformal p-value function.  This pivot is regarded by Vovk as an approximate probability distribution function and is used to define a ``conformal predictive distribution''.  Such a conformal predictive distribution is defined for nonconformity scores such that the conformal p-value function is an increasing function of the to-be-predicted $y$, i.e., also generalizing the Dempster-Hill procedure.  The model-free GF paradigm does not, however, rely on monotonicity of the nonconformity score because it avoids direct construction of a probability distribution function.  This is avoided by explicitly dealing with the imprecise probabilities that arise from the fiducial switching arguments, as discussed from the beginning of Section \ref{sec:imprecision}.  GF and conformal predictive distributions can be regraded as two philosophically different but related interpretations of fiducial predictions \citep[Vovk embraced such a perspective in][]{vovk2024}.

In the context of the Dempster-Hill procedure, the next two results demonstrate the non-asymptotic, sub-exponential concentration that establishes the root-$n$ consistency of all probability measures $\Delta \in \mathscr{C}(\mu)$.  In particular, Theorem \ref{theorem:consistency_1} demonstrates that point-wise, $|\Delta\{(-\infty,y]\} - F(y)| = o_{p}(n^{-\gamma})$, for any $\gamma \in [0,.5)$, where $F$ is the distribution function associated with the true distribution of $Y_{n+1} \sim P$.  Theorem \ref{theorem:consistency_2} establishes an even faster rate of convergence on the focal sets; $|\Delta\{A_{n}(v)\} - P\{A_{n}(v)\}| = o_{p}(n^{-\tau})$, for any $\tau \in [0,1)$.  These results are expected from well-established consistency properties of the empirical distribution function, in the context of empirical process theory.  Namely, the disjoint focal sets $A_{n}(1), \dots, A_{n}(n+1)$ will be narrower and clustered in regions of high probability density, and wider and fewer in regions of low probability density.

\begin{theorem}\label{theorem:consistency_1}
Let $Y_{1}, \dots, Y_{n}\overset{\text{iid}}{\sim} P$ be a collection of continuous random variables and $t_{i}(Y_{i}) := Y_{i}$ for $i \in \{1,\dots,n\}$.  For any $\Delta \in \mathscr{C}(\mu)$, $y \in \R$, $\gamma \in [0,.5)$, $\epsilon > 0$, and $n > 4n^{\gamma}/\epsilon - 1$,
\[
P\Big(n^{\gamma}\big|\Delta\{(-\infty,y]\} - F(y)\big| > \epsilon\Big) \le 4e^{-\frac{\epsilon^2}{8}n^{1-2\gamma}} \cdot 1\{F(y)>0\}.
\]

\end{theorem}
{\noindent \bf Proof.}  Denote $B := (-\infty,y]$.  For any $\Delta \in \mathscr{C}(\mu)$, by definition, $\underline{\mu}(B) \le \Delta(B) \le \overline{\mu}(B)$, and so
\[
|\Delta(B) - P(B)| \le \max\big\{ |\underline{\mu}(B) - P(B)|, |\overline{\mu}(B) - P(B)|\big\}.
\]
A union bound then gives
\[
P\Big(n^{\gamma}\big|\Delta\{(-\infty,y]\} - F(y)\big| > \epsilon\Big) \le P\Big(n^{\gamma}\big|\underline{\mu}(B) - P(B)\big| > \epsilon\Big) + P\Big(n^{\gamma}\big|\overline{\mu}(B) - P(B)\big| > \epsilon\Big),
\]
which suggests that the desired result follows by establishing the stochastic concentration of the belief and plausibility functions around $P(B) = F(y)$.  The latter concentration is established next, next.

Let $M_{n}$ be the number of focal sets having a nonempty intersection with $B$:
\[
M_{n} := \big|\big\{v \, : \, (-\infty,y]\cap A_{n}(v) \ne \emptyset \big\}\big|.
\]
Due to the choice of $t_{i}(Y_{i}) := Y_{i}$ for the nonconformity score, $A_{n}(1) = (-\infty, Y_{(1)}]$, $A_{n}(2) = (Y_{(1)}, Y_{(2)}]$, \dots, $A_{n}(n+1) = (Y_{(n)}, \infty)$, and so
$
M_{n} = 1 + \sum_{i=1}^{n}1\{Y_{i} \le y\}
$
which implies that $Z_{n} := M_{n} - 1 \sim \text{binomial}(n,p_{B})$, where $p_{B} := P(B) = F(y)$.  Then, by definition, $\overline{\mu}(B) = M_{n} / (n+1)$.  Accordingly,
\begin{align}
P\Big(n^{\gamma}\big|\overline{\mu}(B) - P(B)\big| > \epsilon\Big) & = P\Bigg(\bigg|\frac{M_{n}}{n+1} - p_{B}\bigg| > \frac{\epsilon}{n^{\gamma}}\Bigg) \nonumber \\
& \le 1\bigg\{\frac{1}{n+1}  > \frac{\epsilon}{2n^{\gamma}}\bigg\} + P\Bigg(\bigg|\frac{M_{n}-1}{n+1} - p_{B}\bigg| > \frac{\epsilon}{2n^{\gamma}}\Bigg) \nonumber \\
& = P\Bigg(\bigg|\frac{Z_{n}}{n+1} - p_{B}\bigg| > \frac{\epsilon}{2n^{\gamma}}\Bigg) \label{theorem:consistency_1:eq1}, 
\end{align}
where the final equivalence follows by assumption.  If $p_{B} = F(y) = 0$, then all of the above terms are 0 due to the fact that $Z_{n} = 0$ with probability $(1 - p_{B})^{n} = 1$.  Conversely, if $p_{B} = F(y) > 0$, then by the triangle inequality
\begin{align*}
P\Big(n^{\gamma}\big|\overline{\mu}(B) - P(B)\big| > \epsilon\Big) & \le P\bigg(|Z_{n} - n\cdot p_{B}| + p_{B} > \frac{(n+1)\epsilon}{2n^{\gamma}}\bigg) \\
& \le P\bigg(|Z_{n} - E(Z_{n})| > \frac{(n+1)\epsilon}{4n^{\gamma}}\bigg) + 1\bigg\{ p_{B} > \frac{(n+1)\epsilon}{4n^{\gamma}}\bigg\} \\
& \le 2e^{-2\{\frac{(n+1)\epsilon}{4n^{\gamma}}\}^{2} /n} + 0,
\end{align*}
where the last approximation is an application of the Hoeffding inequality (regarding $Z_{n}$ as the sum of $n$ independent Bernoulli$(p_{B})$ random variables), and the assumption that $n > 4n^{\gamma}/\epsilon - 1$.  The more concise expression in the theorem statement employs the fact that $n + 1 \ge n \ge 1$ implies $(n+1)(\frac{n+1}{n}) \ge n + 1 \ge n$, so that
\[
e^{-2\{\frac{(n+1)\epsilon}{4 n^{\gamma}}\}^{2}/n} = e^{-2\frac{\epsilon^{2}}{16 n^{2\gamma}}(n+1)(\frac{n+1}{n})} \le e^{-\frac{\epsilon^2}{8 n^{2\gamma}}n} = e^{-\frac{\epsilon^{2}}{8}n^{1-2\gamma}}.
\]

Next, employ a similar argument to establish the concentration of the belief function:
\[
\big|\big\{v \, : \, A_{n}(v) \subseteq (-\infty,y] \big\}\big| = \sum_{i=1}^{n}1\{Y_{i} \le y\} = Z_{n} \sim \text{binomial}(n,p_{B}),
\]
so that, by definition, $\underline{\mu}(B) = Z_{n}/(n+1)$.  Then
\begin{align*}
P\Big(n^{\gamma}\big|\underline{\mu}(B) - P(B)\big| > \epsilon\Big) & = P\Bigg(\bigg|\frac{Z_{n}}{n+1} - p_{B}\bigg| > \frac{\epsilon}{n^{\gamma}}\Bigg) \le 2e^{-\frac{\epsilon^{2}}{2}n^{1-2\gamma}},
\end{align*}
where the inequality is established using the same arguments following after the final expression in equation (\ref{theorem:consistency_1:eq1}).  Using the fact that $1/2 > 1/8$, it follows that both the belief and plausibility stochastic concentrations can be bound above by $2e^{-\frac{\epsilon^{2}}{8}n^{1-2\gamma}}$.  Adding these two bounds, as indicated by the union bound derived at the beginning of the proof, establishes the theorem statement. \hfill $\blacksquare$\\ 
\begin{theorem}\label{theorem:consistency_2}
Let $Y_{1}, \dots, Y_{n}\overset{\text{iid}}{\sim} P$ be a collection of continuous random variables and $t_{i}(Y_{i}) := Y_{i}$ for $i \in \{1,\dots,n\}$.  For any $\Delta \in \mathscr{C}(\mu)$, $v \in \{1,\dots,n+1\}$, $\tau \in [0,1)$, and $\epsilon > 0$, 
\[
P\Big(n^{\tau}\big|\Delta\{A_{n}(v)\} - P\{A_{n}(v)\}\big| > \epsilon\Big) = 1 - (1-b_{n})^{n} + (1-c_{n})^{n},
\]
where $b_{n} := \max\big\{\frac{1}{n+1} - \frac{\epsilon}{n^{\tau}}, 0\big\}$, $c_{n} := \min\big\{\frac{1}{n+1} + \frac{\epsilon}{n^{\tau}}, 1\big\}$.  In particular, 
\[
1 - (1-b_{n})^{n} + (1-c_{n})^{n} \le e^{-n^{1-\tau}\epsilon},
\] 
for all $n > \max\big\{\frac{n^{\tau} - \epsilon}{\epsilon}, \frac{\epsilon}{n^{\tau}-\epsilon}\big\}$.
\end{theorem}
{\noindent \bf Proof.}  By continuity and independence, denoting $Y_{(0)} := -\infty$ and $Y_{(n+1)} := \infty$, for any $v \in \{1,\dots,n+1\}$,
$
W_{n} := F(Y_{(v)}) - F(Y_{(v-1)}) \sim \text{beta}(1,n).
$
Then, by Lemma \ref{lemma:focalpts},
\begin{align}
P\Big(n^{\tau}\big|\Delta\{A_{n}(v)\} - P\{A_{n}(v)\}\big| > \epsilon\Big) & = P\Bigg(n^{\tau}\bigg|\frac{1}{n+1} - \big[F(Y_{(v)}) - F(Y_{(v-1)})\big]\bigg| > \epsilon\Bigg) \\
& = 1 - P\Bigg(\bigg|W_{n} - \frac{1}{n+1}\bigg| \le \frac{\epsilon}{n^{\tau}}\Bigg) \nonumber \\
& = 1 - P\big(b_{n} \le W_{n} \le c_{n}\big). \nonumber
\end{align}
Next, 
\begin{align*}
P\big(b_{n} \le W_{n} \le c_{n}\big) & = \int_{b_{n}}^{c_{n}}n(1-x)^{n-1} \, dx = - (1-x)^{n} \, \bigg|_{b_{n}}^{c_{n}} = (1-b_{n})^{n} - (1-c_{n})^{n}. 
\end{align*}
Finally, for all $n > \max\big\{\frac{n^{\tau} - \epsilon}{\epsilon}, \frac{\epsilon}{n^{\tau}-\epsilon}\big\}$, it follows that $b_{n} = 0$ and
\[
1 - (1-b_{n})^{n} + (1-c_{n})^{n} = (1-c_{n})^{n} \le e^{-nc_{n}} \le e^{-n^{1-\tau}\epsilon},
\] 
where the first inequality follows by the property of the exponential function that $1 + x \le e^{x}$ for all $x \in \R$.  In particular, set $x = -c_{n}$ and then, noting that $c_{n} \in (0,1)$, take the $n$-th power of both sides.  The second inequality follows from
$
n c_{n} = \frac{n}{n+1} + \frac{n\epsilon}{n^{\tau}} \ge n^{1-\tau} \epsilon.
$
\hfill $\blacksquare$\\ 

\subsection{Connection to the inferential models framework}\label{sec:IM}

A related construction to the model-free GF setup is the {\em nonparametric IM} construction introduced in \cite{cella2022}.  The difference is that, whereas the model-free GF is determined by a mass function over the {\em disjoint} random focal sets $A_{n}(V^{\star})$, the nonparametric IM is determined via a contour function over {\em nested} predictive random sets.  In the notation of Section \ref{sec:mfgf} the nonparametric IM is defined by the contour function $f_{n}(y) = \mu\big\{ \Omega_{n}(V^{\star}) \ni y \big\}$ recalling $\Omega_{n}(\cdot)$ as defined in (\ref{eq:omega}), which is the same as the GF contour function given in (\ref{eq:gf_contour}).  Furthermore, whereas the GF plausibility and belief functions are given by $\overline{\mu}(\cdot)$ and $\underline{\mu}(\cdot)$, respectively, in Section \ref{sec:imprecision}, the IM plausibility and belief functions can be expressed as:
\begin{align*}
\overline{\zeta}(B) & := \sum_{j=1}^{n+1}\mu\{A_{n}(j)\}\cdot 1\big\{\Omega_{n}(j)\cap B \ne \emptyset\big\}, \text{ and} \\
\underline{\zeta}(B) & := \sum_{j=1}^{n+1}\mu\{A_{n}(j)\}\cdot 1\big\{\Omega_{n}(j)\subseteq B\big\}.
\end{align*}

The nested random sets in the IM construction result from the IM being built to satisfy a notion of validity for imprecise probabilities, as stated next in Definition \ref{validity_type2}.
\begin{definition}[Type 2 validity -- \cite{cella2022}]\label{validity_type2}
Let $\overline{\Pi}$ be an upper probability constructed from observed data $y_{1}, \dots, y_{n} \sim Y$, and assume $y_{n+1} \sim Y$.  Denoting by $P$ the probability measure associated with $Y$, the lower and upper probability pair $(\underline{\Pi},\overline{\Pi})$ is type 2 valid if, for all $(\alpha, n, P)$,
$
P\big\{\overline{\Pi}(B) \le \alpha \text{ and } Y_{n+1} \in B \text{ for some } B\big\} \le \alpha.
$
\end{definition}
The utility of an imprecise probability that satisfies type 2 validity is a guarantee that for any event $B$ containing the true value of that to-be-predicted (e.g., corresponding to a null hypothesis), the upper probability (be it a measure of plausibility, possibility, etc.) assigned to $B$ will not be too small too often (over replications of $y_{1}, \dots, y_{n}, y_{n+1}$).  This ensures finite-sample type 1 error protection for general purpose imprecise probabilistic inference, i.e., without pre-specifying or fixing an event/hypothesis $B$.

Mathematically, type 2 validity is achieved by the IM because $\overline{\zeta}(\cdot)$ is a {\em consonant} plausibility function.  A consonant plausibility function is a plausibility function $\overline{\Pi}$ that satisfies $\overline{\Pi}(B) = \sup_{y\in B}f(y)$, for any $B \subseteq \Y$, for some function $f : \Y \to [0,1]$ with $\sup_{y\in \Y}f(y) = 1$.  It is established on page 122 of \cite{cella2022} that $\overline{\zeta}(B) = \sup_{y\in B}f_{n}(y)$, where again, $f_{n}(\cdot)$ also appears as the GF contour function in (\ref{eq:gf_contour}) used to construct the CP sets from the model-free GF distribution.  Therein lies the GF--IM--CP connection.

Notice that $(\underline{\zeta}, \overline{\zeta})$ are constructed here from the GF mass function $\mu(\cdot)$, demonstrating that a consonant plausibility function and thus a type 2 valid imprecise probabilistic predictor is available within the model-free GF context, as well (consonance was also implicit in establishing the GF-CP connection in Theorem \ref{gf_contour_valid}).  Note that it is the belief function $\underline{\zeta}(B) = 1 - \overline{\zeta}(B^{c}) = 1 - \sup_{y\in B^{c}}f_{n}(y)$ that yields $k / (n+1)$ when $B$ is a $k / (n+1)$ level CP set, i.e., when $B = \Omega_{n}(k)$.

Type 2 validity, however, is not satisfied by $(\underline{\mu}, \overline{\mu})$ based on the following argument.  The focal sets $A_{n}(1), \dots, A_{n}(n+1)$ are mutually disjoint and collectively exhaustive, and so for any $y \in \Y$, there exists a single $j \in \{1,\dots,n+1\}$ such that $y \in A_{n}(j)$.  Accordingly, $\overline{\mu}(\{y\}) = \overline{\mu}\{A_{n}(j)\} = 1/(n+1)$, which implies that for any $\alpha \in [1/(n+1), 1)$,
\begin{align*}
\alpha & < 1 \\
& = P\big[\overline{\mu}(\{Y_{n+1}\}) \le \alpha \text{ and } Y_{n+1} \in \{Y_{n+1}\} \big] \\
& \le P\big\{\overline{\mu}(B) \le \alpha \text{ and } Y_{n+1} \in B \text{ for some } B\big\}.
\end{align*}
Thus, while type 2 validity is satisfied for belief and plausibility pairs constructed from $\Omega_{n}(V^{\star})$, i.e., the union of $A_{n}(1), \dots, A_{n}(V^{\star})$, it is not satisfied for belief and plausibility pairs constructed directly from $A_{n}(V^{\star})$.  

There do, nonetheless, remain utilities to imprecise inference based on $(\underline{\mu}, \overline{\mu})$.  First, even though it is not type 2 valid, it does guarantee type 1 error protection for events $B$ that correspond to CP sets, as established by the developments in Section \ref{sec:mfgf}.  Second, consistency, as in Theorems \ref{theorem:consistency_1} and \ref{theorem:consistency_2}, follows more directly from the $(\underline{\mu}, \overline{\mu})$ construction versus the $(\underline{\zeta}, \overline{\zeta})$ construction.  Third, based on the characterization of Lemma \ref{lemma:focalpts}, the $(\underline{\mu}, \overline{\mu})$ construction makes it easy to find pignistic transformations, i.e., the credal set $\mathscr{C}(\mu)$ simply consists of all probability measures assigning $1/(n+1)$ probability over each $A_{n}(v)$.  Such a pignistic transformation is illustrated and others discussed in the following section.  Furthermore, using this illustration, Figures \ref{hist_gaussian} and \ref{hist_mixture} give a visual depiction of the differences between the consonant IM construction $(\underline{\zeta}, \overline{\zeta})$ versus $(\underline{\mu}, \overline{\mu})$.

\section{Illustration of a pignistic transformation}\label{sec:pignistic_illustration}

The preceding developments make a case for the essential role of imprecise probabilistic reasoning in finite-sample, valid inference.  If, however, a decision maker is pressed to make a decision, a transformation from the credal level to the pignistic level might be considered, i.e., the choice of a single probability distribution from the credal set.  Such transformations have been considered previously \citep[e.g., see][]{smets1990,dubois2004}, but no general theory for their construction/evaluation exists \citep{grunwald2018}.  Rather than attempting to propose general principles here, this section serves to simply illustrate how such a transformation can be constructed based on the well-studied concept of a maximum entropy distribution (MED) with respect to Lebesgue measure, for simplicity of the illustration.  The optimality of such a concept in this context is postponed to future work.  Nonetheless, the usual intuition for maximizing entropy is a desire to be least informative or maximally ignorant with respect to unknown information.  

\begin{figure}[H]
\centering
\includegraphics[scale=.26, trim={1.5mm 0mm 15mm 0mm}, clip]{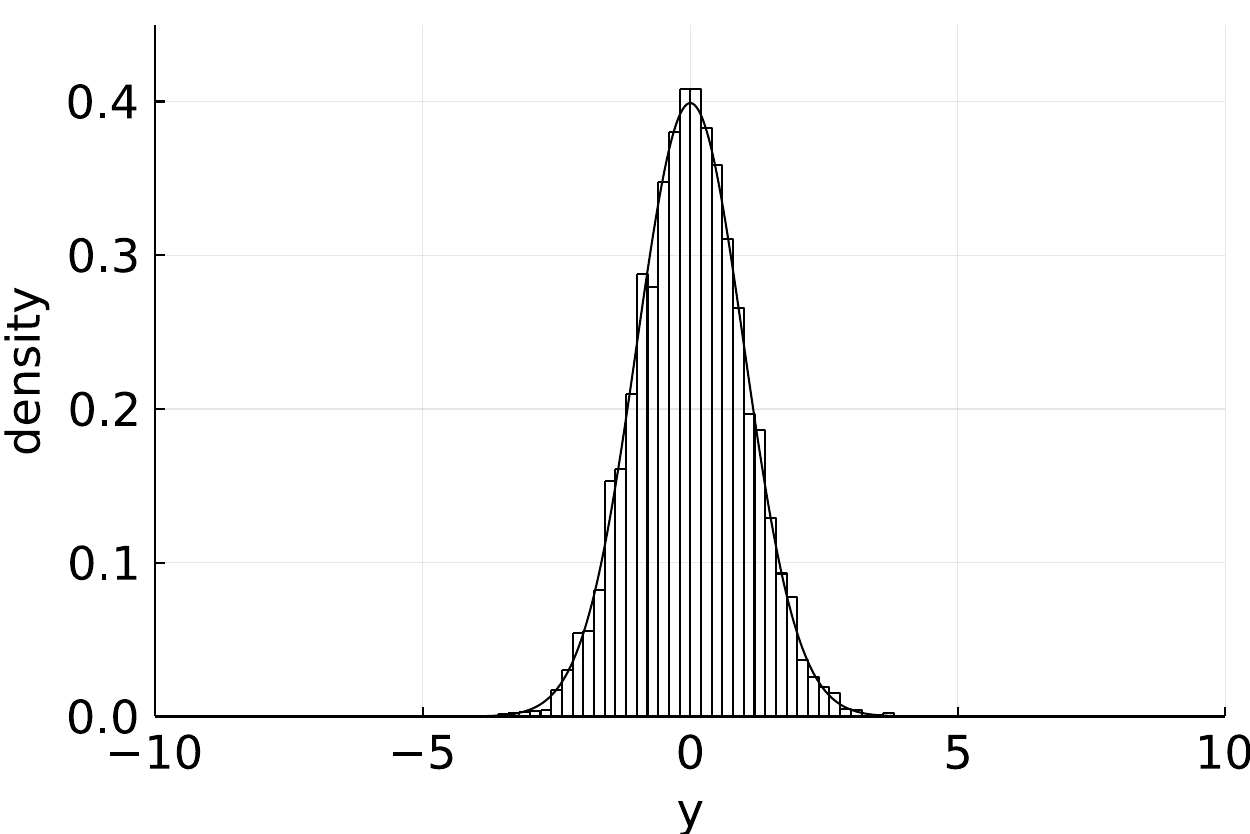}
\includegraphics[scale=.26, trim={11mm 0mm 15mm 0mm}, clip]{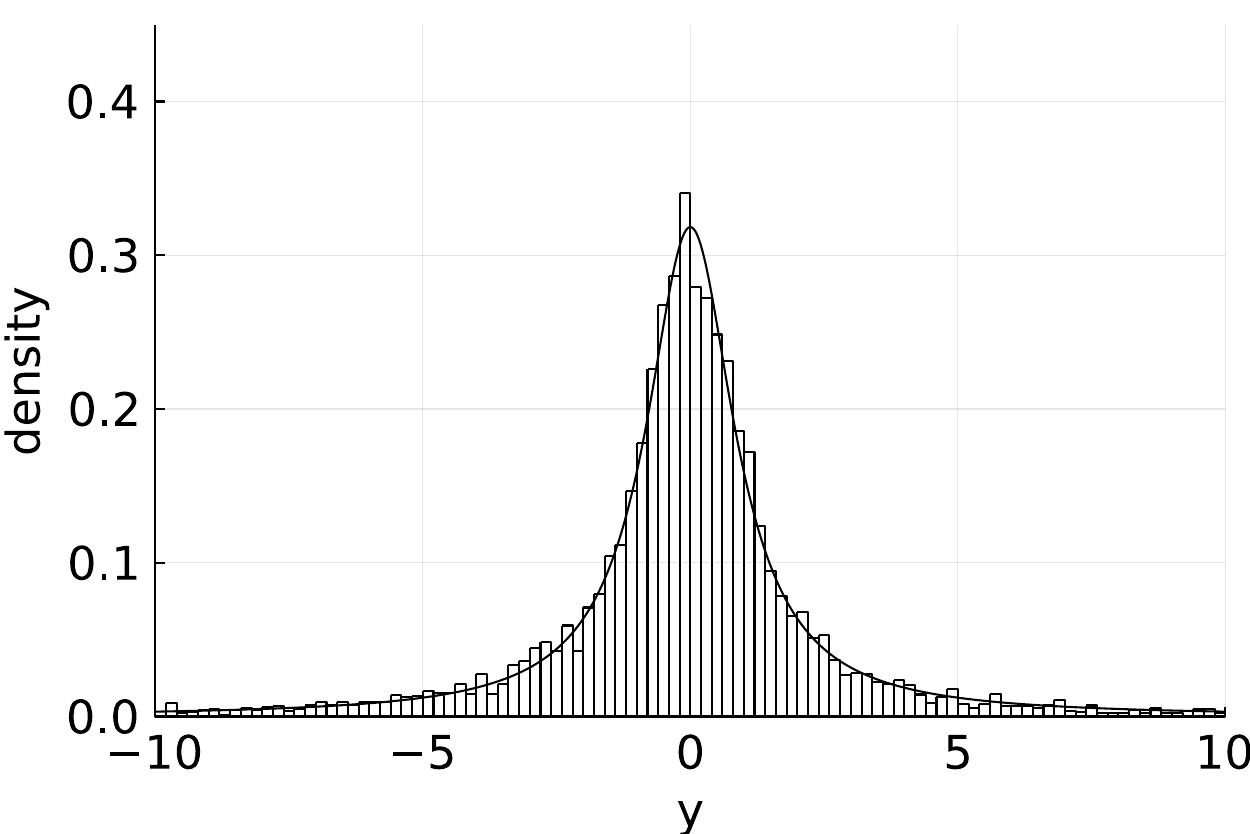}
\includegraphics[scale=.26, trim={11mm 0mm 20mm 0mm}, clip]{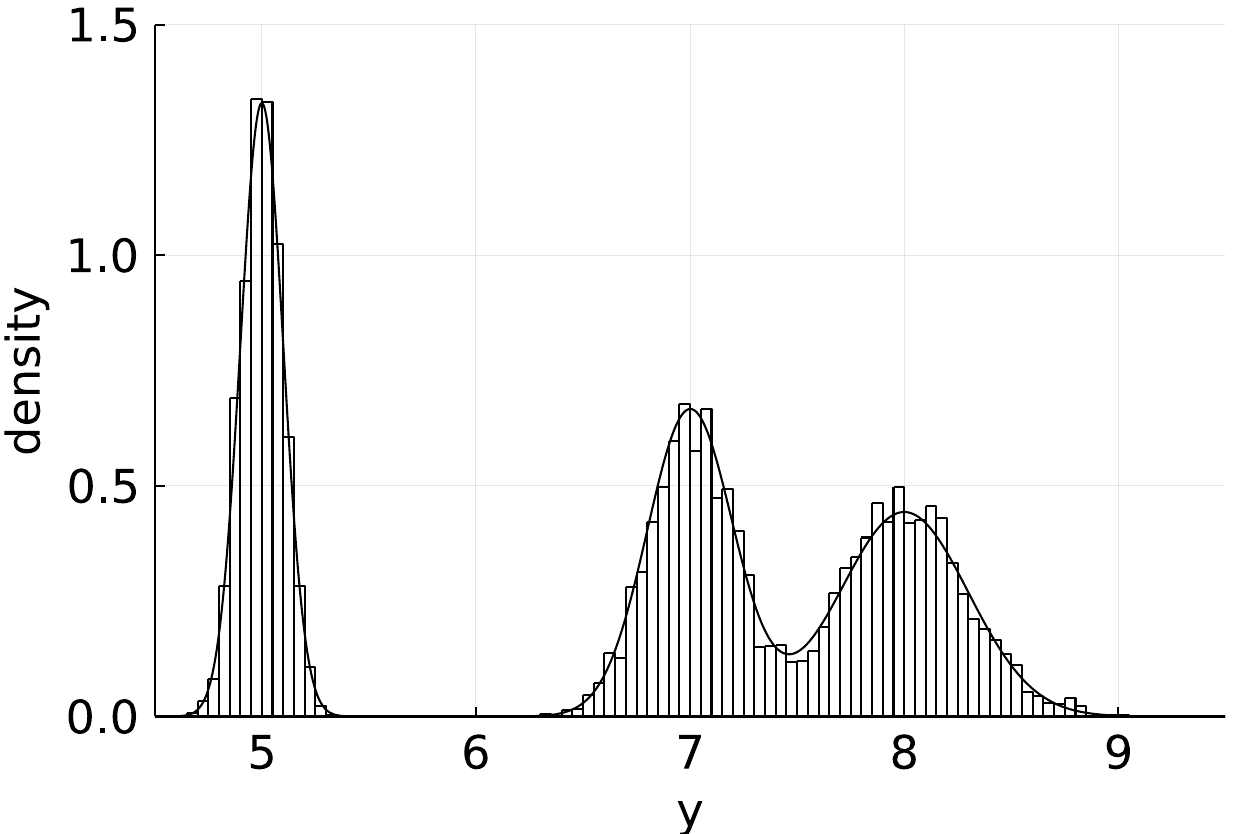}
\caption{\footnotesize Histograms of samples from $\pi_{y}^{n}$ computed by Algorithm \ref{eq:mfgf_alg}, and based on data sets of size $n = 10,000$ from the standard Gaussian distribution (left panel), standard Cauchy distribution (middle panel), and a mixture distribution (right panel).  The nonconformity measure is $t(y_{i}) := y_{i}$.  The black lines are plots of the respective density functions associated with the data.}\label{fig:consistency}
\end{figure}

It is well-known that the MED over a bounded interval is uniform, and so the MED over the credal set $\mathscr{C}(\mu)$ should have a density $\pi^{n}_{y}$ that is flat over each focal set $A_{n}(v)$.  Such a construction might require the modification of $A_{n}(1)$ and/or $A_{n}(n+1)$ so that the support of $\pi^{n}_{y}$ is restricted to $[\kappa_{n}^{\min},\kappa_{n}^{\max}]$ for arbitrarily small/large data-dependent choices of $\kappa_{n}^{\min}$ and $\kappa_{n}^{\max}$, so that uniform densities will integrate over these focal regions.  The MED is derived by integrating the conditional uniform density on every focal set with respect to the GF measure associated with the auxiliary variable: for $y \in [\kappa_{n}^{\min},\kappa_{n}^{\max}]$,
\begin{align}\label{eq:med_density}
\pi^{n}_{y}(y) & = \sum_{v=1}^{n+1} \pi^{n}_{y\mid v}(y \mid v) \cdot \pi^{n}_{v}(v) = \sum_{v=1}^{n+1} 
\begin{cases}
\frac{1}{\lambda\{A_{n}(v)\}} \cdot \frac{1}{n+1} 1\big\{y \in A_{n}(v)\big\} & \text{ if } |A_{n}(v)| > 1 \\
\delta_{A_{n}(v)}(y) \cdot \frac{1}{n+1} & \text{ if } |A_{n}(v)| = 1 \\
\end{cases},
\end{align}
where $\lambda$ is the Lebesgue measure on $\Y$, $|\cdot|$ denotes the cardinality of a set-valued argument, and $\delta_{A_{n}(v)}(\cdot)$ is the Dirac delta function for a singleton set $A_{n}(v)$.  It is established shortly, by Lemma \ref{lemma:credal_set} and Theorem \ref{theorem:med_gf}, that $\pi^{n}_{y}$ is in fact the density associated with the MED over $\mathscr{C}(\mu)$.  Moreover, sampling from this distribution is intuitive, and is described by Algorithm \ref{eq:mfgf_alg}; see Figure \ref{fig:consistency} for an empirical illustration.
\begin{lemma}\label{lemma:credal_set}
The probability measure $\Pi^{n}_{y}(\cdot) := \int_{(\cdot)}\pi^{n}_{y}(y) \, dy \in \mathscr{C}(\mu)$.
\end{lemma}
{\noindent \bf Proof.} For any $j \in \{1,\dots,n+1\}$,
\[
\Pi^{n}_{y}\{A_{n}(j)\} = \int_{A_{n}(j)}\pi^{n}_{y}(y) \, dy =
\begin{cases}
\int_{A_{n}(j)}\frac{1}{\lambda\{A_{n}(j)\}} \cdot \frac{1}{n+1} \, dy = \frac{1}{n+1} & \text{ if } |A_{n}(j)| > 1 \\
\int_{A_{n}(j)}\delta_{A_{n}(j)}(y) \cdot \frac{1}{n+1}  \, dy = \frac{1}{n+1} & \text{ if } |A_{n}(j)| = 1
\end{cases}.
\]
Thus, $\Pi^{n}_{y} \in \mathscr{C}(\mu)$ as a consequence of Lemma \ref{lemma:focalpts}.
\hfill $\blacksquare$\\ 

\medskip
\begin{algorithm}[H]\label{eq:mfgf_alg}\small
\KwInput{Prediction regions $A_{n}(1), \dots, A_{n}(n+1)$.}
\KwOutput{A realized instance of the random variable with density function $\pi^{n}_{y}$.}
Sample $v^{\star} \sim \text{uniform}\{1,\dots,n+1\}$\;
Sample $y^{\star} \sim \text{uniform}\{A_{n}(v^{\star})\}$\;
\Return{$y^{\star}$\;}
\caption{Sampling according to the MED density $\pi^{n}_{y}$ from equation (\ref{eq:med_density}).}
\end{algorithm}
\medskip
\begin{theorem}\label{theorem:med_gf}
The probability distribution having density function $\pi^{n}_{y}$ is the MED over all probability measures in $\mathscr{C}(\mu)$, supported on $[\kappa_{n}^{\min},\kappa_{n}^{\max}]$.
\end{theorem}
{\noindent \bf Proof.} As illustrated by Lemma \ref{lemma:focalpts}, the MED has a density residing in the set of density functions $\mathscr{Q}$ such that for every $q \in \mathscr{Q}$ and for every $v \in \{1,\dots,n+1\}$,
\[
\int_{A_{n}(v)} q(y) \,dy = \frac{1}{n+1}.
\]
If $|A_{n}(v)| = 1$, then $q(y) = \delta_{A_{n}(v)}(y) \cdot \frac{1}{n+1}$ over $A_{n}(v)$.  Alternatively, over the non-singleton focal set regions, the MED over $\mathscr{C}(\mu)$ can be found via the method of Lagrange multipliers constrained to the set $\mathscr{Q}$.  The constrained entropy functional has the form
\[
J[q] = - \int_{\kappa_{n}^{\min}}^{\kappa_{n}^{\max}} q(y)\log\{q(y)\} \, dy + \sum_{j \, : \, |A_{n}(j)| > 1}\beta_{j}\Bigg[\int_{A_{n}(j)}q(y)\,dy - \frac{1}{n+1}\Bigg],
\]
and can be minimized using standard techniques from calculus of variations \citep[a standard text on this subject is][]{gelfand2000}.  The first-order condition for an optimum, based on the functional derivative is
\[
\frac{\delta J}{\delta q} = - \log\{q(y)\} - 1 + \sum_{j \, : \, |A_{n}(j)| > 1}\beta_{j} 1\{y \in A_{n}(j)\} = 0.
\]
Thus, the MED density has the form
$
q(y) = e^{- 1 + \sum_{j \, : \, |A_{n}(j)| > 1}\beta_{j} 1\{y \in A_{n}(j)\}},
$
subject to the constraint
\begin{align*}
\frac{1}{n+1} & = \int_{A_{n}(v)} e^{- 1 + \sum_{j \, : \, |A_{n}(j)| > 1}\beta_{j} 1\{y \in A_{n}(j)\}} \,dy  = \int_{A_{n}(v)} e^{- 1 + \beta_{v}} \,dy = e^{- 1 + \beta_{v}} \lambda\{A_{n}(v)\},
\end{align*}
and so $q(y) = \frac{1}{\lambda\{A_{n}(v)\}} \cdot \frac{1}{n+1}$ for $y \in A_{n}(v)$ for every $v \in \{1,\dots,n+1\}$.  Therefore,
\begin{align*}
q(y) & = \sum_{v=1}^{n+1}
\begin{cases}
\frac{1}{\lambda\{A_{n}(v)\}} \cdot \frac{1}{n+1} 1\big\{y \in A_{n}(v)\big\} & \text{ if } |A_{n}(v)| > 1 \\
\delta_{A_{n}(v)}(y) \cdot \frac{1}{n+1} & \text{ if } |A_{n}(v)| = 1 \\
\end{cases} \\
& = \pi^{n}_{y}(y).
\end{align*}
\hfill $\blacksquare$\\ 

\begin{figure}[H]
\centering
\includegraphics[scale=.5, trim={0mm 0mm 0mm 20mm}, clip]{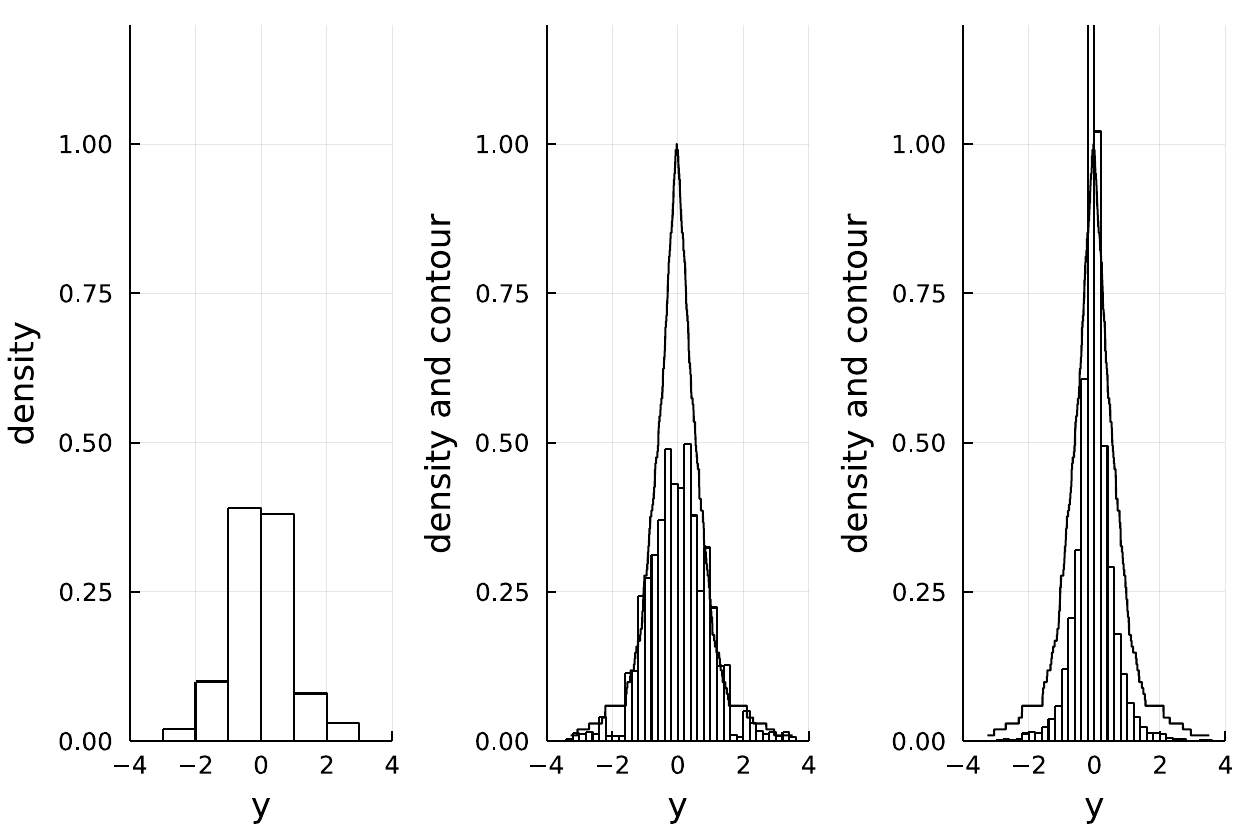}
\caption{\footnotesize Based on the $n = 100$ data points drawn from the standard Gaussian distribution, as summarized by the histogram in the left panel, the middle and right panels display histograms of samples of size 10,000 drawn, respectively, from Algorithm \ref{eq:mfgf_alg} and its analogue by replacing $A_{n}(v^{\star})$ with $\Omega_{n}(v^{\star})$.  The nonconformity measure is $t_{i}(y_{i}) := |\text{mean}(y_{-i}^{n+1}) - y_{i}|$.  For reference, the contour function is provided as the black line in the middle and right panels.}\label{hist_gaussian}
\end{figure}

\begin{figure}[H]
\centering
\includegraphics[scale=.5, trim={0mm 0mm 0mm 30mm}, clip]{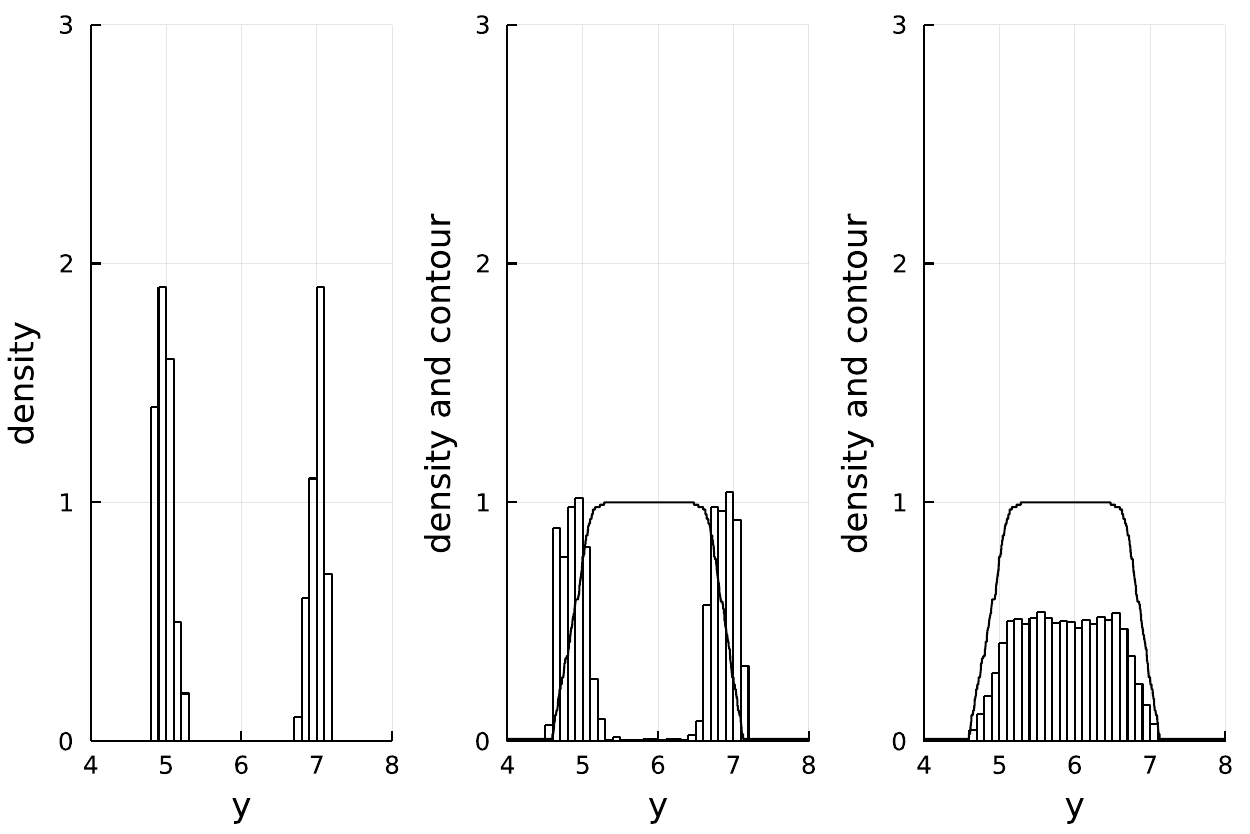}
\caption{\footnotesize Based on the $n = 100$ data points drawn from a mixture of two Gaussian distributions, as summarized by the histogram in the left panel, the middle and right panels display histograms of samples of size 10,000 drawn, respectively, from Algorithm \ref{eq:mfgf_alg} and its analogue by replacing $A_{n}(v^{\star})$ with $\Omega_{n}(v^{\star})$.  The nonconformity measure is $t_{i}(y_{i}) := |\text{mean}(y_{-i}^{n+1}) - y_{i}|$.  For reference, the contour function is provided as the black line in the middle and right panels.}\label{hist_mixture}
\end{figure}

An analogous sampling procedure to Algorithm \ref{eq:mfgf_alg} can be constructed by sampling from $\Omega_{n}(v^{\star})$ rather than $A_{n}(v^{\star})$ in Algorithm \ref{eq:mfgf_alg}.  A comparison of both such algorithms in Figures \ref{hist_gaussian} and \ref{hist_mixture} gives a visual depiction of the differences between the consonant IM construction $(\underline{\zeta}, \overline{\zeta})$ versus $(\underline{\mu}, \overline{\mu})$.  The point of these figures is to highlight the modal tendency of the consonance property.


\subsection{Other pignistic transformations from the GF literature}

Although heuristic methods have been discussed and evaluated empirically \citep{hannig2009}, mapping GF imprecise distributions to precise distributions has remained largely an open question for research in the GF literature.  Recall the example of the imprecise GF distribution constructed for the parameter of a binomial distribution, in Section \ref{sec:gf_intro}.  The existing GF inference literature suggests mapping the imprecise GF distribution to a precise distribution by taking some point in the interval $(U_{(y)}^{\star},U_{(y+1)}^{\star}]$, which can be expressed as suggested in \cite{hannig2009} as
\[
\mathcal{R}_{\theta}(y) = U_{(y)}^{\star} + D(U_{(y+1)}^{\star} - U_{(y)}^{\star}),
\]
for some random variable $D$ supported on or in $[0,1]$.  There are five options for the choice of $D$ discussed in \cite{hannig2009}.  The first is the maximum entropy $D \sim \text{uniform}(0,1)$, and the second is the maximum variance $D \sim \text{uniform}\{0,1\}$ which amounts to arithmetic averaging of the densities of the endpoints.  The third choice is $D \sim \text{beta}(.5,.5)$ which leads to the Bayesian posterior of $\mathcal{R}_{\theta}(y)$ using the Jeffreys prior.  This also corresponds to the geometric mean of the densities of the endpoints, and is advocated for by \cite{schweder2016}.  The fourth choice is
\[
D \mid U_{1}^{\star},\dots,U_{n}^{\star} = 
\begin{cases}
0 & \text{ with probability } U_{(y)}^{\star} \\
1 & \text{ with probability } 1 - U_{(y+1)}^{\star} \\
\text{uniform}(0,1) & \text{ with probability } U_{(y+1)}^{\star} - U_{(y)}^{\star}
\end{cases},
\]
resulting in $\mathcal{R}_{\theta}(y) \sim \text{beta}(y+1,n-y+1)$.  The fifth choice is simply to take the midpoint of the interval (i.e., $D = .5$).  It is observed in simulation studies presented in \cite{hannig2009} that the second choice is optimal in some sense, though, there is a lack of intuition for why it seems to work better than simply taking the midpoint of the interval between the auxiliary endpoints.

\section{Concluding remarks}\label{sec:conclusion}

Systematic investigation of pignistic transformations is critical to the broader understanding and appeal of imprecise probabilistic reasoning from mainstream statistical and machine learning communities, which are mostly familiar only with precise probabilistic reasoning derived from the Kolmogorov axioms.  As stated previously, however, pignistic transformations have not received much attention, even in the imprecise probability literature.  One notable exception is the brief Section 3.2 of \cite{dubois2004} where such transformations are considered as a converse to the main results of the paper, and the proposed transformation relies heavily on a uni-modality assumption.  Another example is \cite{smets1990}, relying on the insufficient reason principle.  There remain many open research questions concerning pignistic transformations, particularly from a statistical inference perspective.  Exploring optimization strategies---as motivated in this article by Theorem \ref{theorem:med_gf}---in a new research area of {\em calculus of variations on credal sets} could be fruitful for new constructions of credal sets and pignistic transformations guided by finite-sample frequentist validity properties.

\acks{This manuscript benefited substantially from numerous discussions and discourse with Ryan Martin, Peter Gr{\"u}nwald, and one anonymous reviewer.}


\newpage
\appendix
\section{}


{\noindent\bf Proof of Theorem \ref{theorem:cp_cons_valid}.}  Any realization of nonconformity scores $t_{1}(y_{1}),\dots,t_{n+1}(y_{n+1})$ can be described as a collection containing unique values, $a_{1},\dots,a_{K}$, for some $K \le n+1$ and occurring with frequencies $n_{1},\dots,n_{K}$, respectively (with $\sum_{k=1}^{K}n_{k} = n+1$).  Using the fact that $t_{1}(Y_{1}),\dots,t_{n+1}(Y_{n+1})$ are exchangeable (because $Y_{1},\dots,Y_{n+1}$ are exchangeable) as in Definition \ref{exchangeability}, it follows by definition that any realization $t_{1}(y_{1}),\dots,t_{n+1}(y_{n+1})$ can be understood as some permutation of the values in the {\em bag} (i.e., a collection of elements with no ordering),
\[
B := \lbag \underbrace{a_{(1)}, \dots, a_{(1)}}_{n_{1}}, \underbrace{a_{(2)}, \dots, a_{(2)}}_{n_{2}}, \dots, \underbrace{a_{(K)}, \dots, a_{(K)}}_{n_{K}} \rbag,
\]
where $a_{(k)}$ is the $k$-th order statistic (in ascending order) of the values $a_{1},\dots,a_{K}$.  As such, the observed nonconformity scores $t_{1}(y_{1}),\dots,t_{n+1}(y_{n+1})$ are just one of $(n+1)!$ equally possible permutations that could have been recorded, assuming $y_{n+1}$ was generated from equation (\ref{dge}).  

Next, with reference to the bag $B$ it can be determined that
\[
\sum_{i=1}^{n+1}1\{t_{i}(y_{i}) \ge t_{n+1}(y_{n+1})\} =
\begin{cases}
n+1 & \text{ if } \ t_{n+1}(y_{n+1}) = a_{(1)} \\
n+1-n_{1}  & \text{ if } \ t_{n+1}(y_{n+1}) = a_{(2)} \\
n+1-n_{1}-n_{2} & \text{ if } \ t_{n+1}(y_{n+1}) = a_{(3)} \\
& \vdots \\
n_{K} & \text{if } \ t_{n+1}(y_{n+1}) = a_{(K)}
\end{cases}.
\]
Furthermore, there are $n!\cdot n_{j}$ permutations of the values in $B$ in which the last reported value, $t_{n+1}(y_{n+1}) = a_{(j)}$, so it must be the case that
\[
P\bigg(\sum_{i=1}^{n+1}1\{t_{i}(Y_{i}) \ge t_{n+1}(Y_{n+1})\} = v \ \Big| \ B\bigg) = 
\begin{cases}
\frac{n!\cdot n_{1}}{(n+1)!} = \frac{n_{1}}{n+1} & \text{ if } \ v = n+1 \\
\frac{n!\cdot n_{2}}{(n+1)!} = \frac{n_{2}}{n+1} & \text{ if } \ v = n+1-n_{1} \\
\frac{n!\cdot n_{3}}{(n+1)!} = \frac{n_{3}}{n+1} & \text{ if } \ v = n+1-n_{1}-n_{2} \\
& \vdots \\
\frac{n!\cdot n_{K}}{(n+1)!} = \frac{n_{K}}{n+1} & \text{ if } \ v = n_{K} \\
0 & \text{ else } 
\end{cases}.
\]
Note that in the special case without repeated values (i.e., $n_{1} = \cdots = n_{K} = 1$), the above expression reduces to a discrete uniform probability mass function.  In any case,
\begin{align*}
P( \Gamma_{n}^{\alpha} \not\ni Y_{n+1} \mid B) & = P\{p_{n}(Y_{n+1}) \le \alpha \mid B\} \\
& = P\bigg(\sum_{i=1}^{n+1}1\{t_{i}(Y_{i}) \ge t_{n+1}(Y_{n+1})\} \le \alpha (n+1) \ \Big| \ B\bigg) \\
& = 
\begin{cases}
\frac{n_{K}}{n+1} +  \frac{n_{K-1}}{n+1} + \cdots + \frac{n_{k_{\alpha}+1}}{n+1} & \text{ if } k_{\alpha} < K \\
0 & \text{ if } k_{\alpha} = K
\end{cases},
\end{align*}
where $k_{\alpha} := \min\big\{j \in \{0,\dots,K\} \, : \, n+1-\sum_{k=0}^{j}n_{k} \le \alpha (n+1)\big\}$ and $n_{0} := 0$.  Observe from the construction of $k_{\alpha}$ that, 
\[
n_{K} + n_{K-1} + \cdots + n_{k_{\alpha}+1} = n+1-\sum_{k=0}^{k_{\alpha}}n_{k} \le \alpha (n+1),
\]
and divide by $n+1$ on all sides.  Thus, in any case,
\[
P( \Gamma_{n}^{\alpha} \not\ni Y_{n+1}) = \int P( \Gamma_{n}^{\alpha} \not\ni Y_{n+1} \mid B) d\nu(B) \le \alpha \cdot \int d\nu(B) = \alpha,
\]
where $\nu(\cdot)$ is any probability measure that described the uncertainty in observing the bag $B$.
\hfill $\blacksquare$

\vskip 0.2in
\bibliography{/Users/jwilli27/Documents/Research/references.bib}

\begin{thebibliography}{49}
\providecommand{\natexlab}[1]{#1}
\providecommand{\url}[1]{\texttt{#1}}
\expandafter\ifx\csname urlstyle\endcsname\relax
  \providecommand{\doi}[1]{doi: #1}\else
  \providecommand{\doi}{doi: \begingroup \urlstyle{rm}\Url}\fi

\bibitem[Angelopoulos et~al.(2023)Angelopoulos, Bates, Fisch, Lei, and
  Schuster]{angelopoulos2023}
A.~N. Angelopoulos, S.~Bates, A.~Fisch, L.~Lei, and T.~Schuster.
\newblock Conformal risk control.
\newblock In \emph{The Twelfth International Conference on Learning
  Representations}, 2023.

\bibitem[Angelopoulos et~al.(2025)Angelopoulos, Bates, Cand{\`e}s, Jordan, and
  Lei]{angelopoulos2025}
A.~N. Angelopoulos, S.~Bates, E.~J. Cand{\`e}s, M.~I. Jordan, and L.~Lei.
\newblock Learn then test: {C}alibrating predictive algorithms to achieve risk
  control.
\newblock \emph{Annals of Applied Statistics}, 19\penalty0 (2):\penalty0
  1641--1662, 2025.

\bibitem[Augustin and Coolen(2004)]{augustin2004}
T.~Augustin and F.~P. Coolen.
\newblock Nonparametric predictive inference and interval probability.
\newblock \emph{Journal of Statistical Planning and Inference}, 124\penalty0
  (2):\penalty0 251--272, 2004.

\bibitem[Balch et~al.(2019)Balch, Martin, and Ferson]{balch2019}
M.~S. Balch, R.~Martin, and S.~Ferson.
\newblock Satellite conjunction analysis and the false confidence theorem.
\newblock \emph{Proceedings of the Royal Society A}, 475\penalty0 (20180565),
  2019.

\bibitem[Basir and Yuan(2007)]{basir2007}
O.~Basir and X.~Yuan.
\newblock Engine fault diagnosis based on multi-sensor information fusion using
  {D}empster-{S}hafer evidence theory.
\newblock \emph{Information fusion}, 8\penalty0 (4):\penalty0 379--386, 2007.

\bibitem[Bates et~al.(2021)Bates, Angelopoulos, Lei, Malik, and
  Jordan]{bates2021}
S.~Bates, A.~Angelopoulos, L.~Lei, J.~Malik, and M.~Jordan.
\newblock Distribution-free, risk-controlling prediction sets.
\newblock \emph{Journal of the ACM (JACM)}, 68\penalty0 (6):\penalty0 1--34,
  2021.

\bibitem[Bloch(1996)]{bloch1996}
I.~Bloch.
\newblock Some aspects of {D}empster-{S}hafer evidence theory for
  classification of multi-modality medical images taking partial volume effect
  into account.
\newblock \emph{Pattern Recognition Letters}, 17\penalty0 (8):\penalty0
  905--919, 1996.

\bibitem[Carmichael and Williams(2018)]{carmichael2018}
I.~Carmichael and J.~P. Williams.
\newblock An exposition of the false confidence theorem.
\newblock \emph{Stat}, 7\penalty0 (1):\penalty0 e201, 2018.

\bibitem[Cella and Martin(2022)]{cella2022}
L.~Cella and R.~Martin.
\newblock Validity, consonant plausibility measures, and conformal prediction.
\newblock \emph{International Journal of Approximate Reasoning}, 141:\penalty0
  110--130, 2022.

\bibitem[Coolen(1998)]{coolen1998}
F.~P. Coolen.
\newblock Low structure imprecise predictive inference for {B}ayes' problem.
\newblock \emph{Statistics \& Probability Letters}, 36\penalty0 (4):\penalty0
  349--357, 1998.

\bibitem[Coolen and Coolen-Maturi(2024)]{coolen2024}
F.~P. Coolen and T.~Coolen-Maturi.
\newblock Nonparametric predictive inference.
\newblock In \emph{International Encyclopedia of Statistical Science (2nd
  Edition)}. Springer, 2024.

\bibitem[Dempster(1963)]{dempster1963}
A.~P. Dempster.
\newblock On direct probabilities.
\newblock \emph{Journal of the Royal Statistical Society: Series B},
  25\penalty0 (1):\penalty0 100--110, 1963.

\bibitem[Dempster(1966)]{dempster1966}
A.~P. Dempster.
\newblock New methods for reasoning towards posterior distributions based on
  sample data.
\newblock \emph{Annals of Mathematical Statistics}, 37\penalty0 (2):\penalty0
  355--374, 1966.

\bibitem[Denoeux(2000)]{denoeux2000}
T.~Denoeux.
\newblock A neural network classifier based on {D}empster-{S}hafer theory.
\newblock \emph{IEEE Transactions on Systems, Man, and Cybernetics-Part A:
  Systems and Humans}, 30\penalty0 (2):\penalty0 131--150, 2000.

\bibitem[Denoeux(2008)]{denoeux2008}
T.~Denoeux.
\newblock A k-nearest neighbor classification rule based on {D}empster-{S}hafer
  theory.
\newblock In \emph{Classic works of the Dempster-Shafer theory of belief
  functions}, pages 737--760. Springer, 2008.

\bibitem[Dey and Williams(2023)]{dey2023}
N.~Dey and J.~P. Williams.
\newblock Valid inference for machine learning model parameters.
\newblock \emph{arXiv preprint arXiv:2302.10840}, 2023.

\bibitem[Dey et~al.(2025)Dey, Martin, and Williams]{dey2024}
N.~Dey, R.~Martin, and J.~P. Williams.
\newblock Anytime-valid generalized universal inference on risk minimizers.
\newblock \emph{Journal of the Royal Statistical Society: Series B}, 2025.

\bibitem[D{\'\i}az-M{\'a}s et~al.(2010)D{\'\i}az-M{\'a}s, Mu{\~n}oz-Salinas,
  Madrid-Cuevas, and Medina-Carnicer]{diaz-mas2010}
L.~D{\'\i}az-M{\'a}s, R.~Mu{\~n}oz-Salinas, F.~J. Madrid-Cuevas, and
  R.~Medina-Carnicer.
\newblock Shape from silhouette using {D}empster-{S}hafer theory.
\newblock \emph{Pattern Recognition}, 43\penalty0 (6):\penalty0 2119--2131,
  2010.

\bibitem[Dubois et~al.(2004)Dubois, Foulloy, Mauris, and Prade]{dubois2004}
D.~Dubois, L.~Foulloy, G.~Mauris, and H.~Prade.
\newblock Probability-possibility transformations, triangular fuzzy sets, and
  probabilistic inequalities.
\newblock \emph{Reliable computing}, 10\penalty0 (4):\penalty0 273--297, 2004.

\bibitem[Gelfand and Fomin(2000)]{gelfand2000}
I.~Gelfand and S.~Fomin.
\newblock Calculus of variations,(translated and edited by silverman, ra),
  2000.

\bibitem[Gr{\"u}nwald(2018)]{grunwald2018}
P.~Gr{\"u}nwald.
\newblock Safe probability.
\newblock \emph{Journal of Statistical Planning and Inference}, 195:\penalty0
  47--63, 2018.

\bibitem[Gr{\"u}nwald(2024)]{grunwald2024a}
P.~Gr{\"u}nwald.
\newblock Beyond {N}eyman--{P}earson: {E}-values enable hypothesis testing with
  a data-driven alpha.
\newblock \emph{Proceedings of the National Academy of Sciences}, 121\penalty0
  (39):\penalty0 e2302098121, 2024.

\bibitem[Gr{\"u}nwald et~al.(2024)Gr{\"u}nwald, de~Heide, and
  Koolen]{grunwald2024}
P.~Gr{\"u}nwald, R.~de~Heide, and W.~Koolen.
\newblock Safe testing.
\newblock \emph{Journal of the Royal Statistical Society: Series B},
  86\penalty0 (5):\penalty0 1091--1128, 2024.

\bibitem[Hannig(2009)]{hannig2009}
J.~Hannig.
\newblock On generalized fiducial inference.
\newblock \emph{Statistica Sinica}, 19\penalty0 (2):\penalty0 491--544, 2009.

\bibitem[Hannig et~al.(2016)Hannig, Iyer, Lai, and Lee]{hannig2016}
J.~Hannig, H.~Iyer, R.~C. Lai, and T.~C. Lee.
\newblock Generalized fiducial inference: A review and new results.
\newblock \emph{Journal of the American Statistical Association}, 111\penalty0
  (515):\penalty0 1346--1361, 2016.

\bibitem[Hill(1968)]{hill1968}
B.~M. Hill.
\newblock Posterior distribution of percentiles: {B}ayes' theorem for sampling
  from a population.
\newblock \emph{Journal of the American Statistical Association}, 63\penalty0
  (322):\penalty0 677--691, 1968.

\bibitem[Hill(1988)]{hill1988}
B.~M. Hill.
\newblock {De Finetti’s theorem, induction, and A(n) or Bayesian
  nonparametric predictive inference (with discussion)}.
\newblock \emph{Bayesian statistics}, 3:\penalty0 211--241, 1988.

\bibitem[Hjort et~al.(2025)Hjort, Hermansen, Pensar, and Williams]{hjort2025}
A.~Hjort, G.~H. Hermansen, J.~Pensar, and J.~P. Williams.
\newblock Uncertainty quantification in automated valuation models with
  spatially weighted conformal prediction.
\newblock \emph{International Journal of Data Science and Analytics}, pages
  1--18, 2025.

\bibitem[Jacob et~al.(2021)Jacob, Gong, Edlefsen, and Dempster]{jacob2021}
P.~E. Jacob, R.~Gong, P.~T. Edlefsen, and A.~P. Dempster.
\newblock A {G}ibbs sampler for a class of random convex polytopes.
\newblock \emph{Journal of the American Statistical Association}, 116\penalty0
  (535):\penalty0 1181--1192, 2021.

\bibitem[Jeffreys(1932)]{jeffreys1932}
H.~Jeffreys.
\newblock On the theory of errors and least squares.
\newblock \emph{Proceedings of the Royal Society of London. Series A,
  Containing Papers of a Mathematical and Physical Character}, 138\penalty0
  (834):\penalty0 48--55, 1932.

\bibitem[Martin(2019)]{martin2019}
R.~Martin.
\newblock False confidence, non-additive beliefs, and valid statistical
  inference.
\newblock \emph{International Journal of Approximate Reasoning}, 113:\penalty0
  39--73, 2019.

\bibitem[Martin and Liu(2015)]{martin2015}
R.~Martin and C.~Liu.
\newblock \emph{Inferential models: reasoning with uncertainty}, volume 145.
\newblock CRC Press, 2015.

\bibitem[Park et~al.(2023)Park, Balakrishnan, and Wasserman]{park2023}
B.~Park, S.~Balakrishnan, and L.~Wasserman.
\newblock Robust universal inference.
\newblock \emph{arXiv preprint arXiv:2307.04034}, 2023.

\bibitem[Ramdas et~al.(2023)Ramdas, Gr{\"u}nwald, Vovk, and Shafer]{ramdas2023}
A.~Ramdas, P.~Gr{\"u}nwald, V.~Vovk, and G.~Shafer.
\newblock Game-theoretic statistics and safe anytime-valid inference.
\newblock \emph{Statistical Science}, 38\penalty0 (4):\penalty0 576--601, 2023.

\bibitem[Schweder and Hjort(2016)]{schweder2016}
T.~Schweder and N.~L. Hjort.
\newblock \emph{Confidence, likelihood, probability}, volume~41.
\newblock Cambridge University Press, 2016.

\bibitem[Seidenfeld and Wasserman(1993)]{seidenfeld1993}
T.~Seidenfeld and L.~Wasserman.
\newblock Dilation for sets of probabilities.
\newblock \emph{Annals of Statistics}, 21\penalty0 (3):\penalty0 1139--1154,
  1993.

\bibitem[Shafer(1976)]{shafer1976}
G.~Shafer.
\newblock \emph{A mathematical theory of evidence}.
\newblock Princeton university press, 1976.

\bibitem[Shafer(2021)]{shafer2021}
G.~Shafer.
\newblock Comment on “{A} {G}ibbs sampler for a class of random convex
  polytopes,” by {P}ierre {E}. {J}acob, {R}uobin {G}ong, {P}aul {T}.
  {E}dlefsen, and {A}rthur {P}. {D}empster.
\newblock \emph{Journal of the American Statistical Association}, 116\penalty0
  (535):\penalty0 1196--1197, 2021.

\bibitem[Smets(1990)]{smets1990}
P.~Smets.
\newblock Constructing the pignistic probability function in a context of
  uncertainty.
\newblock In M.~Henrion, R.~D. Shachter, L.~N. Kanal, and J.~F. Lemmer,
  editors, \emph{Uncertainty in Artificial Intelligence}, volume~10 of
  \emph{Machine Intelligence and Pattern Recognition}, pages 29--39.
  North-Holland, 1990.

\bibitem[Tie et~al.(2022)Tie, Cohen, Lahouel, Lo, Wang, Kosmider, Wong,
  Shapiro, Lee, Harris, Khattak, Burge, Harris, Lynam, Nott, Day, Hayes,
  McLachlan, Lee, Ptak, Silliman, Dobbyn, Popoli, Hruban, Lennon, Papadopoulos,
  Kinzler, Vogelstein, Tomasetti, and Gibbs]{tie2022}
J.~Tie, J.~Cohen, K.~Lahouel, S.~Lo, Y.~Wang, S.~Kosmider, R.~Wong, J.~Shapiro,
  M.~Lee, S.~Harris, A.~Khattak, M.~Burge, M.~Harris, J.~Lynam, L.~Nott,
  F.~Day, T.~Hayes, S.~McLachlan, B.~Lee, J.~Ptak, N.~Silliman, L.~Dobbyn,
  M.~Popoli, R.~Hruban, A.~Lennon, N.~Papadopoulos, K.~Kinzler, B.~Vogelstein,
  C.~Tomasetti, and P.~Gibbs.
\newblock Circulating tumor {DNA} analysis guiding adjuvant therapy in stage
  {II} colon cancer.
\newblock \emph{New England Journal of Medicine}, 386\penalty0 (24):\penalty0
  2261--2272, 2022.

\bibitem[Vasseur et~al.(1999)Vasseur, P{\'e}gard, Mouaddib, and
  Delahoche]{vasseur1999}
P.~Vasseur, C.~P{\'e}gard, E.~Mouaddib, and L.~Delahoche.
\newblock Perceptual organization approach based on {D}empster-{S}hafer theory.
\newblock \emph{Pattern recognition}, 32\penalty0 (8):\penalty0 1449--1462,
  1999.

\bibitem[Vovk(2022)]{vovk2022}
V.~Vovk.
\newblock Universal predictive systems.
\newblock \emph{Pattern Recognition}, 126:\penalty0 108536, 2022.

\bibitem[Vovk(2024)]{vovk2024}
V.~Vovk.
\newblock Conformal predictive distributions: An approach to nonparametric
  fiducial prediction.
\newblock In \emph{Handbook of Bayesian, Fiducial, and Frequentist Inference},
  pages 364--380. Chapman and Hall/CRC, 2024.

\bibitem[Vovk et~al.(2005)Vovk, Gammerman, and Shafer]{vovk2005}
V.~Vovk, A.~Gammerman, and G.~Shafer.
\newblock \emph{Algorithmic learning in a random world}.
\newblock Springer Science \& Business Media, 2005.

\bibitem[Vovk et~al.(2019)Vovk, Shen, Manokhin, and Xie]{vovk2019}
V.~Vovk, J.~Shen, V.~Manokhin, and M.~Xie.
\newblock Nonparametric predictive distributions based on conformal prediction.
\newblock \emph{Machine Learning}, 108:\penalty0 445--474, 2019.

\bibitem[Wasserman et~al.(2020)Wasserman, Ramdas, and
  Balakrishnan]{wasserman2020}
L.~Wasserman, A.~Ramdas, and S.~Balakrishnan.
\newblock Universal inference.
\newblock \emph{Proceedings of the National Academy of Sciences}, 117\penalty0
  (29):\penalty0 16880--16890, 2020.

\bibitem[Williams(2021)]{williams2021}
J.~P. Williams.
\newblock Discussion of “{A} {G}ibbs sampler for a class of random convex
  polytopes”.
\newblock \emph{Journal of the American Statistical Association}, 116\penalty0
  (535):\penalty0 1198--1200, 2021.

\bibitem[Williams and Liu(2024)]{williams2024decision}
J.~P. Williams and Y.~Liu.
\newblock Decision theory via model-free generalized fiducial inference.
\newblock In \emph{Belief Functions: Theory and Applications}, volume 14909,
  pages 131--139. Springer, 2024.

\bibitem[Xie and Singh(2013)]{xie2013}
M.~Xie and K.~Singh.
\newblock Confidence distribution, the frequentist distribution estimator of a
  parameter: A review.
\newblock \emph{International Statistical Review}, 81\penalty0 (1):\penalty0
  3--39, 2013.

\end{thebibliography}

\end{document}